\documentclass[runningheads]{llncs}

\usepackage{graphics} 
\usepackage{graphicx} 
\usepackage{float} 
\usepackage{caption}
\usepackage{subcaption}
\usepackage{graphicx}
\usepackage{float}
\usepackage{amsmath}
\usepackage{amssymb}
\usepackage{url}
\allowdisplaybreaks
\begin{document}
\title{Iterated Invariant EKF for 3D Landmark-Aided Inertial Navigation
\thanks{This study was financed in part by the Coordena\c{c}\~{a}o de Aperfei\c{c}oamento de Pessoal de N\'{\i}vel Superior -- Brasil (CAPES) -- Finance Code 001 and the Funda\c{c}\~{a}o de Amparo \`{a} Pesquisa do Estado do Rio de Janeiro (FAPERJ).}}
%
%
\author{Hilton Marques Souza Santana \inst{1}\orcidID{
0000-0002-2840-2904
} \and
João Carlos Virgolino Soares\inst{2}\orcidID{0000-0002-6278-378X} \and
Marco Antonio Meggiolaro \inst{1}\orcidID{0000-0002-6240-8189}}
\authorrunning{H. Santana et al.}
%
\institute{Pontifical Catholic University of Rio de Janeiro, Rio de Janeiro, Brazil\\
\email{hiltonmarques@gmail.com} \and
Dynamic Legged Systems Lab, Istituto Italiano di Tecnologia, Genova, Italy}
\maketitle              
\begin{abstract}
Inertial navigation systems aided by three-dimensional landmark measurements constitute a fundamental problem in robotic perception and state estimation. Classical $\mathrm{SO}(3)$-based Extended Kalman Filter ($\mathrm{SO}(3)$-EKF) approaches provide practical solutions, but suffer from the false observability problem, in which the filter becomes overconfident in unobservable directions, leading to degraded estimation performance. The Invariant EKF (IEKF) addresses this limitation by reformulating the system dynamics as a group-affine system on a Lie group, although its measurement update does not fully satisfy certain state compatibility properties. More recently, the Iterated Invariant EKF (IterIEKF) was proposed to further improve the IEKF by ensuring, in the low-noise regime, that the estimated state remains on the observed state manifold while the uncertainty is confined to its tangent space. In this work, we formulate and apply the IterIEKF to landmark-based inertial 3D localization for the first time. Through numerical simulations, we show that the proposed approach outperforms the classical $\mathrm{SO}(3)$-EKF, the Iterated $\mathrm{SO}(3)$-EKF, and the IEKF in terms of both estimation accuracy and consistency.

\keywords{Kalman Filter  \and Localization \and Sensor Fusion.}
\end{abstract}
\section{Introduction}
The necessity of estimating the state of a mobile system, including its position, orientation, and velocity, has been a fundamental challenge throughout the history of navigation. The use of celestial landmarks for localization spans several millennia. In particular, mariners in the Northern Hemisphere estimated their latitude by measuring the elevation of the North Star (Polaris) above the horizon~\cite[pg. 2]{barfoot2024}. These early navigation techniques relied on identifiable, static features in the environment, giving rise to the modern notion of landmarks as fixed reference points that can be repeatedly observed to infer an agent's state. During the mid-twentieth century, the success of the Apollo missions demonstrated that robust navigation required not only observations of external landmarks, such as stars, but also accurate inertial measurements from accelerometers and gyroscopes. The fusion of inertial and exteroceptive sensing established many of the principles that underpin modern state estimation.
Since the early 2000s, the emergence of Micro Aerial Vehicles (MAVs), together with the widespread availability of sophisticated sensing technologies, including RGB-D cameras, LiDARs, and high-performance embedded computing platforms, has enabled real-time state estimation and map reconstruction in increasingly complex environments~\cite{elhousni2022}. These technological advances have made it practical to address the Simultaneous Localization and Mapping problem~\cite{dissanayake2001}.

The classical algorithmic foundation for landmark-based localization for autonomous mobile robots was established in~\cite{leonard1991}. 
This framework relies on two key stages: landmark detection within the scene, followed by data association (or matching), which evaluates whether an observed feature has been previously cataloged or represents a new landmark. In this work, we consider a 3D MAV navigating with inertial sensors through an environment where a prior map has already been constructed and features are uniquely identifiable, meaning the underlying data association problem is assumed to be solved. Then, we combine exteroceptive updates (measurements of known landmarks) and proprioceptive inputs (inertial measurements), also called Visual-Inertial Odometry (VIO), through a unified probabilistic framework in order to obtain a real-time state estimate of the robot.

Probabilistic filtering methods, particularly those based on the Extended Kalman Filter (EKF), have played a central role in landmark-aided inertial navigation over the past decades.
One of its earliest applications to inertial navigation with a vehicle equipped with an Inertial Measurement Unit (IMU) and capable of autonomously measuring its relative position with respect to a pre-built three-dimensional map was presented in~\cite{trawny2007}. That work formulated an $\mathrm{SO}(3)$-EKF (also known as Multiplicative-EKF or Quaternion-EKF~\cite{markley2003}) for the precision landing of a spacecraft using observations of known planetary landmarks. Subsequently, Li and Mourikis~\cite{li2013} showed that this standard formulation suffers from the so-called false observability problem, motivating the development of constrained filters. Later, a seminal study by Barrau and Bonnabel~\cite{barrau2016} demonstrated that false observability fundamentally arises from the choice of error-state parameterization and can be mitigated by exploiting the symmetry of the underlying system through the Invariant Extended Kalman Filter (IEKF). The IEKF was later applied to indoor MAV navigation in~\cite{brossard2018}, where it demonstrated superior estimation performance compared to conventional EKF formulations. Nevertheless, the classical IEKF still relies on a first-order linearization of the measurement model about the predicted state, which may degrade estimation accuracy in highly nonlinear scenarios. To overcome this limitation, Goffin et al.~\cite{goffin2026} proposed the Iterated IEKF (IterIEKF), which performs a sequence of local optimization steps to estimate the mode of the posterior distribution while reducing the impact of first-order approximation errors. The IterIEKF has recently demonstrated improved convergence properties for state estimation of a quadrupedal robot in~\cite{santana2026}.

In this work, we apply the IterIEKF framework for the first time to a MAV equipped with an IMU capable of measuring its relative position to known environmental landmarks. Through extensive Monte Carlo simulations evaluated on one of the EuRoC MAV datasets~\cite{burri2016}, we demonstrate that the IterIEKF outperforms the standard $\mathrm{SO}(3)$-EKF, its iterated variant (Iter$\mathrm{SO}(3)$-EKF), and the standard IEKF in terms of state estimation accuracy and statistical consistency. More specifically, the primary contributions of this paper are summarized as follows:
\begin{itemize}
    \item We derive complete mathematical formulations of both invariant and multiplicative iterated filters tailored for landmark-aided MAV state estimation. To the best of our knowledge, the complete algorithmic derivation of the IterIEKF in this specific framework is presented here for the first time.
    
    \item We provide a systematic comparative study analyzing the direct impact of iterative invariant updates on both the trajectory accuracy and the statistical consistency of the estimated states. Through controlled Monte Carlo trials, we demonstrate that the IterIEKF yields substantial improvements in consistency metrics, resulting in an error reduction of approximately $80\%$ in the estimated position and linear velocity vectors.
    
    \item We present an in-depth analysis of online IMU bias estimation dynamics, demonstrating that while the accelerometer bias remains difficult to fully observe across all evaluated filters, the IterIEKF consistently achieves the most accurate profile overall.
\end{itemize}

The remainder of this work is structured as follows. Section~\ref{sec:literature_review} reviews relevant literature, and Section~\ref{sec:mathematical_background} establishes the mathematical background. Section~\ref{sec:motivating_example} provides a geometric motivating example for IterIEKF. Sections~\ref{sec:discrete_time_inertial_model} to~\ref{sec:multiplicative} detail the derivations of the IEKF, IterIEKF, $\mathrm{SO}(3)$-EKF, and Iter$\mathrm{SO}(3)$-EKF filters using inertial and 3D landmark measurements. Section~\ref{sec:results} presents Monte Carlo experimental results evaluating filter accuracy and consistency, and Section~\ref{sec:conclusion} concludes the paper.

\section{Related Work}
\label{sec:literature_review}
The landscape of visual-inertial state estimation has evolved significantly, shifting from traditional local coordinate representations to geometrically principled invariant frameworks and high-fidelity iterative updates. 

\subsection{Multiplicative and Invariant Visual-Inertial Estimators}
Early frameworks for landmark-aided localization primarily focused on absolute positioning and landing applications. For instance, 
Trawny et al.~\cite{trawny2007} developed a vision-aided inertial navigation system specifically optimized for pin-point landing, relying on observations of pre-mapped landmarks. While highly effective for localized target tracking, classical $\mathrm{SO}(3)$-EKF frameworks are fundamentally hindered by consistency issues. This limitation was thoroughly dissected by Li and Mourikis~\cite{li2013}, who demonstrated that conventional VIO formulations suffer from false observability, where the filter erroneously gains information along unobservable directions, such as yaw and global position, leading to overoptimistic covariance estimates and degraded accuracy.

To resolve these consistency mismatches, Barrau and Bonnabel introduced the IEKF framework \cite{barrau2016}. 
Later, Brossard et al.~\cite{brossard2018} applied it directly to visual-inertial SLAM. By parameterizing the state error using Lie group actions, the IEKF guarantees that the error kinematics remain independent of the robot's absolute trajectory. This mathematical property inherently eliminates the false observability problem without requiring artificial modifications like First-Estimates Jacobians. Our work builds directly upon this invariant paradigm, extending it from a single-step propagation-update loop to an optimization-driven iterative routine tailored for MAV setups.

\subsection{Iterated and Manifold-Based Filtering}
Parallel to the development of invariant filtering, iterated Kalman filters emerged as an effective approach to reduce the linearization errors introduced by highly nonlinear measurement models. In the context of visual-inertial odometry, Bloesch et al.~\cite{bloesch2017} proposed an iterated EKF that combines direct photometric feedback with manifold-based state representations for rotations and landmark bearing vectors, improving robustness through iterative measurement refinement. Later, He et al.~\cite{he2021} generalized this methodology into a canonical framework for Kalman filtering on differentiable manifolds, enabling generic iterated error-state filters on arbitrary product manifolds. However, neither approach explicitly addresses the consistency issues arising from the incorrect linearization of unobservable directions, and therefore, they do not resolve the false-observability problem.

Bridging the gap between geometric tracking and invariant filtering, Goffin et al.~\cite{goffin2026} introduced the Iterated IEKF. By fusing Lie group invariance with manifold-aware Gauss-Newton iterations, the IterIEKF solves the maximum a posteriori (MAP) estimation problem while confining uncertainty in the tangent space of the observed submanifold. While their initial framework demonstrated remarkable performance on legged quadrupedal platforms~\cite{santana2026}, its application to unconstrained, high-velocity 3D aerial robotics has yet to be fully explored. 

In contrast to the aforementioned literature, this paper presents the first explicit application and evaluation of the IterIEKF for a 3D MAV executing navigation loops via relative position observations to mapped landmarks. Unlike traditional VIO methods~\cite{trawny2007,li2013} that suffer from false observability, and conventional iterated manifold filters~\cite{bloesch2017,he2021} that lack invariant error dynamics, our formulation combines the statistical consistency of invariant mechanics with the high-accuracy tracking of iterated update steps. We provide a rigorous, simulation-backed verification of this approach under extensive Monte Carlo trials, explicitly detailing the filter's superiority in state accuracy, consistency, and online IMU bias estimation.

\section{Mathematical Background}
\label{sec:mathematical_background}
We define the robot true state at instant $t_i$ as being the orientation 
($\mathbf{R}_i \in  \mathrm{SO}(3)$), base velocity ($\mathbf{v}_i \in  \mathbb{R}^{3}$), position ($\mathbf{p}_i \in \mathbb{R}^{3}$), all relative to the world frame, and IMU gyroscope and 
accelerometer bias ($\mathbf{b}_{g}, \mathbf{b}_a \in  \mathbb{R}^{3}$). In this work, the robot state is embedded into two distinct matrix Lie groups, 
$\mathrm{SE}_2(3) \times  \mathbb{R}^{6}$~\cite{brossard2018} and $\mathrm{SO}(3) \times  \mathbb{R}^{12}$~\cite{trawny2007}:
\begin{equation}
\label{eq:state}
\begin{aligned}
\mathcal{X}_{i} := \begin{bmatrix}
\begin{bmatrix}
\mathbf{R}_{i} & \mathbf{v}_{i} & \mathbf{p}_{i} \\
\mathbf{0}_{1,3}  &  1  & 0 \\
\mathbf{0}_{1,3} & 0 & 1
\end{bmatrix} & \mathbf{0}_{5,7} \\
\mathbf{0}_{7,5} & \begin{bmatrix}
\mathbf{I}_{6} & \begin{bmatrix} \mathbf{b}_{g,i} \\ \mathbf{b}_{a,i} \end{bmatrix} \\
\mathbf{0}_{1,6} & 1
\end{bmatrix}
\end{bmatrix} \in \mathrm{SE}_2(3) \times  \mathbb{R}^{6}, \\
\boldsymbol{x}_i  := \{\mathbf{R}_i, \mathbf{v}_i, \mathbf{p}_i, \mathbf{b}_{g,i}, \mathbf{b}_{a,i}\} \in \mathrm{SO}(3) \times  \mathbb{R}^{12}
.\end{aligned}
\end{equation}
Every matrix Lie group $G$ has an associated Lie algebra $\mathfrak{g}$ of the same dimension $n$. We define the isomorphisms between $\mathfrak{g}$ and $\mathbb{R}^{n}$ as:
\begin{gather}
    \text{Hat}: \mathbb{R}^{n} \to \mathfrak{g}, \quad \mathbf{v} \mapsto \mathbf{v}^{\wedge} = \sum_{i=1}^{n}v_{i}E_{i}, \\
    \text{Vee}: \mathfrak{g} \to \mathbb{R}^{n}, \quad \mathbf{v}^{\wedge} \mapsto (\mathbf{v}^{\wedge})^{\vee} = \mathbf{v} = \sum_{i=1}^{n} v_{i}\mathbf{e}_{i},
\end{gather}
where $\{E_i\}$ and $\{ \mathbf{e}_{i}\}$ are bases for $\mathfrak{g}$ and $\mathbb{R}^{n}$. In this work, we use the $\mathbb{R}^{n}$ representation of $\mathfrak{g}$. Both spaces are related by the local homeomorphism $\exp_G: \mathfrak{g} \to G$, where $\mathbf{v}^{\wedge} \mapsto \exp_G(\mathbf{v}^{\wedge}) = \sum_{i=0}^{\infty} \frac{{\mathbf{v}^{\wedge}}^{i}}{i!}$, and its inverse $\log_G: G \to \mathfrak{g}$. We define $\mathrm{Exp}_G(\mathbf{v}) := \exp_G(\mathbf{v}^{\wedge})$ and $\mathrm{Log}_G(\mathcal{X}) := \log_G(\mathcal{X})^{\vee}$ for $\mathbf{v} \in \mathbb{R}^n, \mathcal{X} \in G$. Following \cite{sola2018}, we overload the left- and right-$\oplus$ operations, 
\begin{equation}
\label{eq:main_operators}
\begin{aligned}
	\text{left-}\oplus: G \times \mathbb{R}^{n} \to  G,\mathcal{X} \oplus \mathbf{u} &= \mathcal{X} \mathrm{Exp}(\mathbf{u}), \\
	\text{right-}\oplus: \mathbb{R}^{n} \times G \to G, \mathbf{u} \oplus \mathcal{X} &= \mathrm{Exp}(\mathbf{u})\mathcal{X},
\end{aligned}
\end{equation}
and approximate the exponential map for $\| \mathbf{u}\| \approx 0$ using the right-Jacobian $\mathcal{J}_{r,G}(\cdot)$:
\begin{equation}
\label{eq:right_jacobian}
\mathrm{Exp}_G(\mathbf{a} + \mathbf{u}) \approx \mathrm{Exp}_G(\mathbf{a}) \oplus \mathcal{J}_{r,G}(\mathbf{a}) \mathbf{u}.
\end{equation}
The Lie algebra is important because it accommodates the error-state vectors of the respective filters~\cite{barrau2022}. In this work, we adopt a right-invariant error parameterization, defining the error-states as zero-mean Gaussian random variables as follows:
\begin{equation}
    \label{eq:error_states}
    \begin{aligned}
        \boldsymbol{\xi}_i &:= \mathrm{Log}_{\mathrm{SE}_2(3)\times \mathbb{R}^{6}}(\mathcal{X}_i \bar{\mathcal{X}}_i^{-1}), \quad \boldsymbol{\xi}_i \sim \mathcal{N}(\mathbf{0}_{15,1}, \mathbf{P}_i),\\ 
        \delta \boldsymbol{x}_i &:= \mathrm{Log}_{\mathrm{SO}(3)\times \mathbb{R}^{12}}(\boldsymbol{x}_i \bar{\boldsymbol{x}}_i^{-1}), \quad \delta \boldsymbol{x}_i \sim \mathcal{N}(\mathbf{0}_{15,1}, \mathbf{\Sigma}_i).
    \end{aligned}
\end{equation}
The exponential, logarithmic, and right-Jacobians for the isolated Lie groups $\mathrm{SE}_2(3)$ and $\mathrm{SO}(3)$ are detailed in ~\cite[Appendix I]{santana2026}. 
Their algebraic extension to the Cartesian product with $\mathbb{R}^{k}$ vector spaces is straightforwardly obtained by forming block-diagonal matrices composed of the original group operators and identity matrices of matching dimensions.
\section{Motivating Example}
\label{sec:motivating_example}

\begin{figure}[h!]
    \centering
    \includegraphics[width=0.28\textwidth]{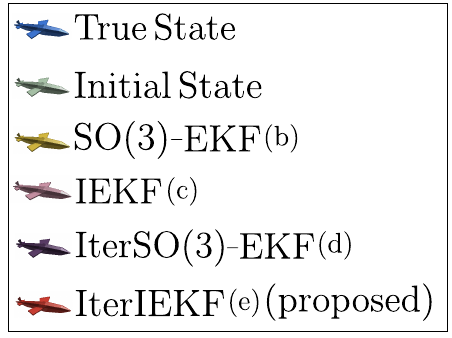}%
    \hfill
    \subfloat[\protect\label{fig:setup_slam}]{\includegraphics[width=0.33\textwidth]{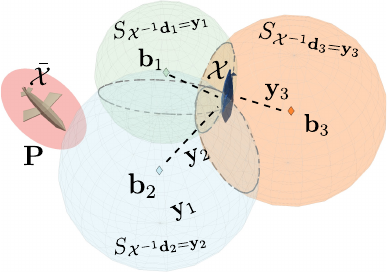}}%
    \hfill
    \subfloat[\protect\label{fig:so3rn3ekf_slam}]{\includegraphics[width=0.33\textwidth]{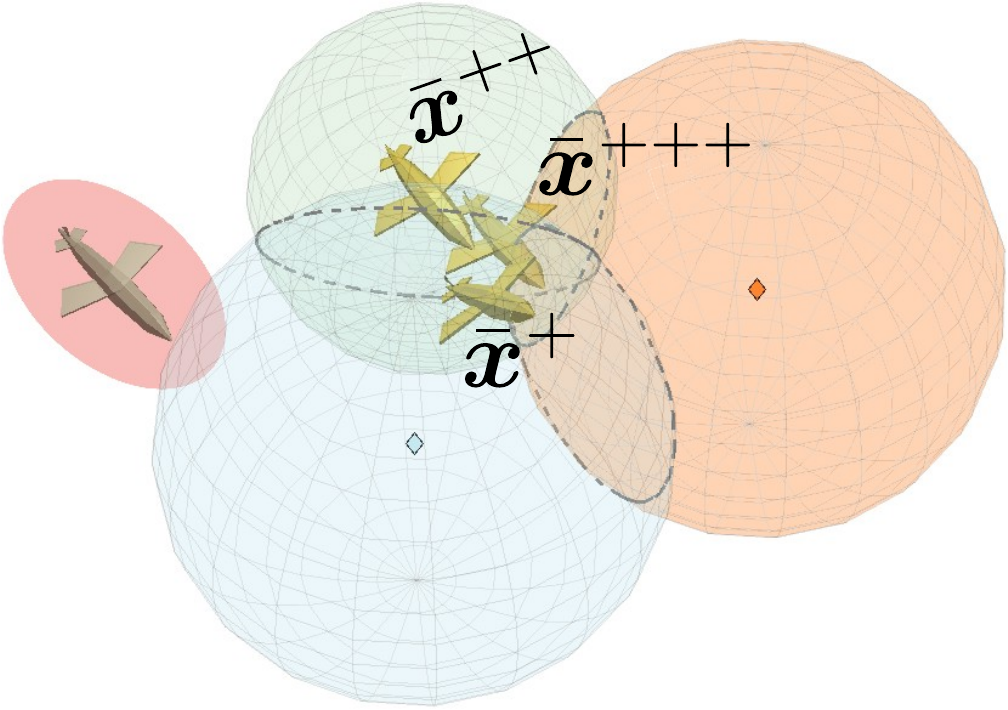}}\\
    \subfloat[\protect\label{fig:iekf_slam}]{\includegraphics[width=0.33\textwidth]{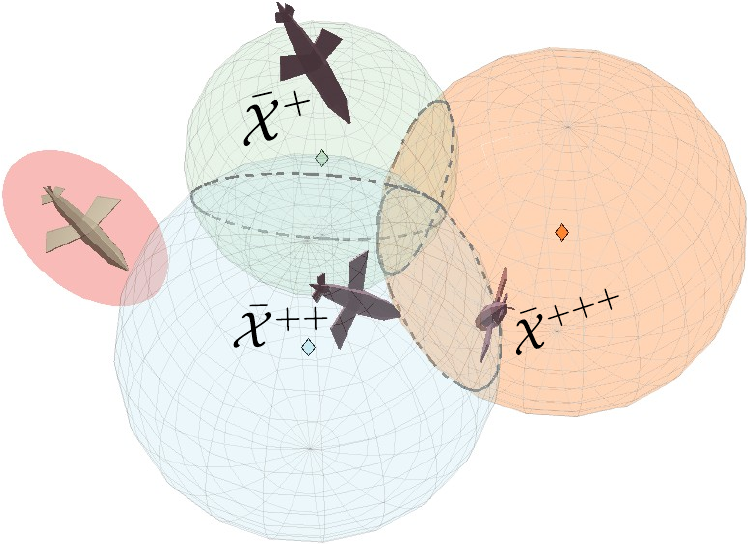}}%
    \hfill
    \subfloat[\protect\label{fig:iterso3r3nekf}]{\includegraphics[width=0.33\textwidth]{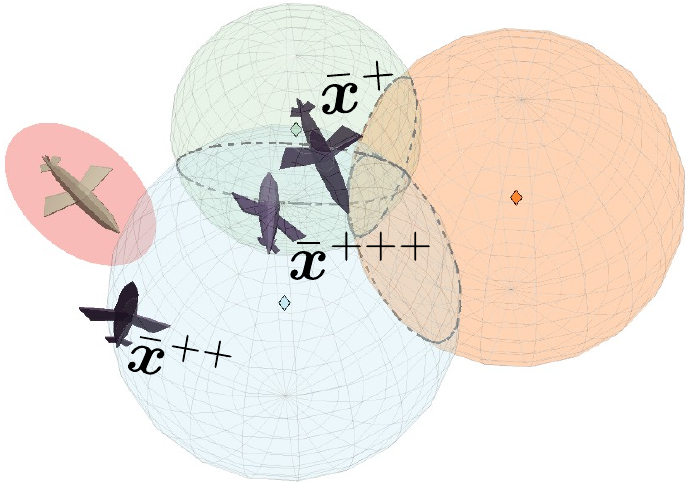}}%
    \hfill
    \subfloat[\protect\label{fig:iteriekf_slam}]{\includegraphics[width=0.33\textwidth]{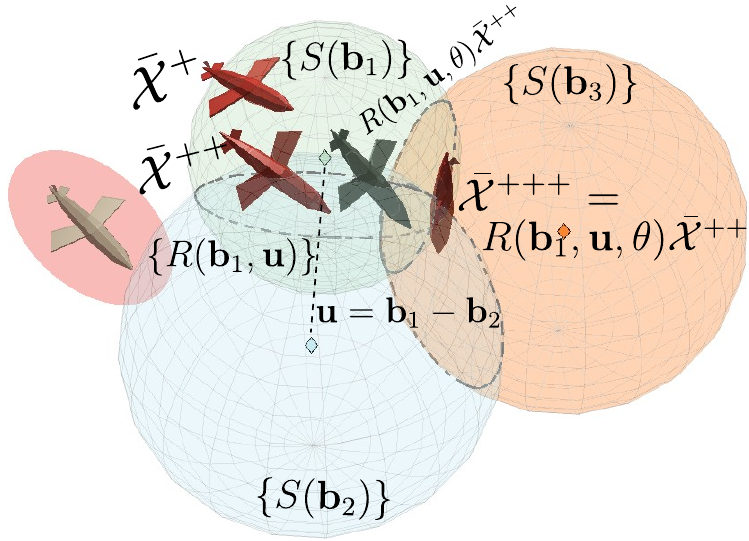}}
    \caption{Consecutive updates with noise-free measurements: a) Problem setup. b) $\text{SO}(3)$-EKF: updates miss the observed sets. c) IEKF and d) Iter$\text{SO}(3)$-EKF: only the first update lands on its set. e) IterIEKF: all updates remain on the observed sets, converging perfectly to their intersection.}
    \label{fig:slam}
\end{figure}

\begin{figure}[h!]
    \centering
    \includegraphics[width=0.40\textwidth]{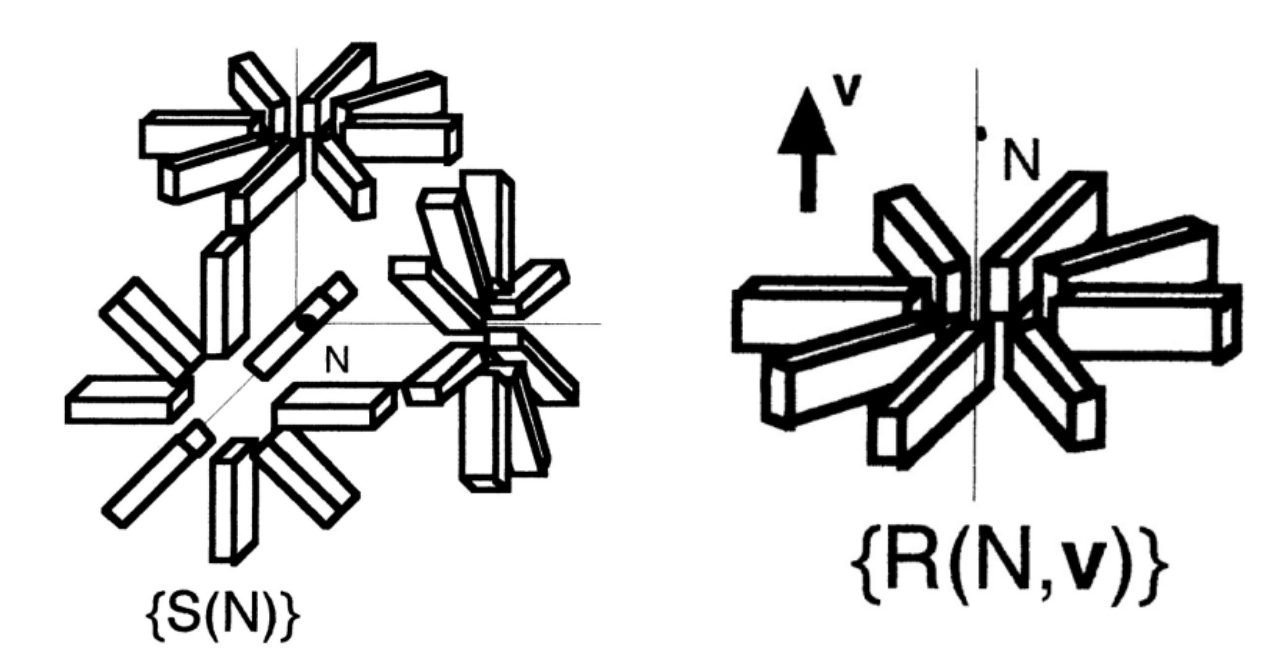}
    \caption{Two stabilizer subgroups of $\text{SE}(3)$: $\{S(N)\} \cong \text{SO}(3)$ and $\{ R(N, \mathbf{v}) \} \cong \text{SO}(2)$~\cite{herve1999}.}
    \label{fig:subgroups_herve}
\end{figure}

To motivate the proposed IterIEKF formulation, we present a 3D localization example with an MAV and three landmarks under nearly noise-free conditions (Fig.~\ref{fig:slam}), extending the 2D case in~\cite[Section IV]{santana2026}. Let $\mathcal{X}$ be the true MAV state and $\bar{\mathcal{X}}$ be the initial nominal state with covariance $\mathbf{P}$, positioned far from $\mathcal{X}$. The robot measures its relative pose to three landmarks, $\mathbf{b}_1, \mathbf{b}_2, \mathbf{b}_3$. Each measurement $k$ defines an observed submanifold $S_{\mathcal{X}^{-1}\mathbf{d_k}=\mathbf{y}_k}$ where the robot could be located~\cite[Definition 1]{santana2026} (Fig.~\ref{fig:setup_slam}). Geometrically, each observed set is a conjugate subgroup $\{S(\mathbf{b}_k)\} \cong \text{SO}(3)$~\cite{herve1999} (Fig.~\ref{fig:subgroups_herve}).
We perform consecutive updates using four filters: $\text{SO}(3)$-EKF, IEKF, Iter$\text{SO}(3)$-EKF, and IterIEKF (Figs.~\ref{fig:so3rn3ekf_slam}--\ref{fig:iteriekf_slam}). As shown, $\text{SO}(3)$-EKF misses all observed sets, while IEKF and Iter$\text{SO}(3)$-EKF converge only to the first. Conversely, IterIEKF successfully reaches the intersection of all three submanifolds, recovering the true state. 

This superior performance stems from the core property of IterIEKF: each update projects the estimate onto the observed manifold while confining uncertainty to its tangent space~\cite[Theorem 1]{goffin2026}. Thus, if the filter converges, the final state lies exactly on the intersection of the submanifolds~\cite[Theorem 2]{goffin2026}. Geometrically, the intersection of the first two subgroups, $\{S(\mathbf{b}_1)\}$ and $\{S(\mathbf{b}_2)\}$, yields a subgroup $\{R(\mathbf{b}_1, \mathbf{u})\} \cong \text{SO}(2)$, where $\mathbf{u} = \mathbf{b}_2 - \mathbf{b}_1$~\cite{herve1999} (Fig.~\ref{fig:subgroups_herve}). After two updates, the state is constrained to $\{R(\mathbf{b}_1,\mathbf{u})\}$, leaving a single degree of freedom. The final update acts on $\bar{\mathcal{X}}^{++}$ within this subgroup to uniquely determine the robot's position and orientation (Fig.~\ref{fig:iteriekf_slam}).

\section{Discrete-Time Inertial Model}
\label{sec:discrete_time_inertial_model}

We assume the system dynamics is governed by IMU measurements, an old approach known as strapdown navigation~\cite{edward1971}.
In this approach, the high-frequency dynamics evolution is obtained by integrating the IMU measurements, i.e., the angular velocity $\tilde{\boldsymbol{\omega}}_{\mathcal{I},i}$ and linear acceleration $\tilde{\mathbf{a}}_{\mathcal{I},i}$. We denote by the subscript $\mathcal{I}$ all quantities expressed in the IMU frame. We assume the measurements are piecewise constant over the time interval $[t_i, t_i + \Delta t]$ and that they are corrupted by additive white Gaussian noise and biases modeled as zero-mean random walks:
\begin{align}
    \tilde{\boldsymbol{\omega}}_{\mathcal{I},i} &= \boldsymbol{\omega}_{\mathcal{I},i} + \mathbf{w}_{g,i} + \mathbf{b}_{g,i}, & \mathbf{w}_{g,i} &\sim \mathcal{N}(\mathbf{0}_{3,1}, \mathbf{Q}_g), \\
    \mathbf{b}_{g,i + 1} &= \mathbf{b}_{g,i} + \mathbf{w}_{bg,i}\Delta t, & \mathbf{w}_{bg,i} &\sim \mathcal{N}(\mathbf{0}_{3,1}, \mathbf{Q}_{bg}), \\
    \tilde{\mathbf{a}}_{\mathcal{I},i} &= \mathbf{a}_{\mathcal{I},i} + \mathbf{w}_{a,i} + \mathbf{b}_{a,i}, & \mathbf{w}_{a,i} &\sim \mathcal{N}(\mathbf{0}_{3,1}, \mathbf{Q}_a), \\
    \mathbf{b}_{a,i + 1} &= \mathbf{b}_{a,i} + \mathbf{w}_{ba,i}\Delta t, & \mathbf{w}_{ba,i} &\sim \mathcal{N}(\mathbf{0}_{3,1}, \mathbf{Q}_{ba}).
\end{align}
In this work, the nominal state is propagated through the system dynamics, while the error state is propagated through the linearized dynamics. 
The error state propagation is discussed in Section~\ref{sec:invariant} for the invariant filters and Section~\ref{sec:multiplicative} for the multiplicative filters.
The system dynamics for all filters considered in this work are governed by the IMU inputs and given by~\cite{sola2017,santana2024}:
\begin{equation}
\label{eq:prediction_step}
\begin{aligned}
\bar{\boldsymbol{x}}_{i|i-1} &:= \mathcal{F}_{i-1}(\bar{\boldsymbol{x}}_{i-1}, \tilde{\boldsymbol{\omega}}_{\mathcal{I},i-1}, \tilde{\mathbf{a}}_{\mathcal{I},i-1}, \Delta t) \\
&\approx \begin{bmatrix}
    \bar{\mathbf{R}}_{i-1} \oplus (\tilde{\boldsymbol{\omega}}_{\mathcal{I},i-1} - \bar{\mathbf{b}}_{g, i-1})\Delta t \\
    \bar{\mathbf{v}}_{i-1} + [\bar{\mathbf{R}}_{i-1}(\tilde{\mathbf{a}}_{\mathcal{I},i-1} - \bar{\mathbf{b}}_{a,i-1}) + \mathbf{g}] \Delta t \\
    \bar{\mathbf{p}}_{i-1} + \bar{\mathbf{v}}_{i-1}\Delta t +  [\bar{\mathbf{R}}_{i-1}(\tilde{\mathbf{a}}_{\mathcal{I},i-1} - \bar{\mathbf{b}}_{a,i-1}) + \mathbf{g}]\frac{\Delta t^{2}}{2} \\
\bar{\mathbf{b}}_{g,i-1}\\
\bar{\mathbf{b}}_{a,i-1}
\end{bmatrix},
\end{aligned}
\end{equation}
where $\mathbf{g} := [0,0,-9.81]^{T}$ and the nominal biases are kept constant during the propagation step. Notice that Eq. \eqref{eq:prediction_step} is also used to obtain $\overline{\mathcal{X}}_{i|i-1}$~\cite{brossard2022}.
\section{3D Landmark Measurement Model}
Let $\mathbf{b}_k$ represent a known landmark in the scene where the MAV is navigating. The MAV can measure the relative position of the landmark $\mathbf{b}_k$ at instant $t_i$ with respect to its IMU frame $\mathcal{I}$, which is denoted as $\mathbf{y}_{k,i}$. This measurement model can be expressed as follows:
\begin{equation}
\label{eq:so3_meas}
\mathbf{y}_{k,i} = \mathbf{R}_i^{T}(\mathbf{b}_k - \mathbf{p}_i) + \mathbf{n}_{k,i}, \quad 
\mathbf{n}_{k,i} \sim \mathcal{N}(\mathbf{0}_{3,1}, \mathbf{N}_{k,i}),
\end{equation}
or alternatively through the group action of $\mathrm{SE}_2(3) \times \mathbb{R}^{6}$ required by invariant filter formulations~\cite{barrau2022}:
\begin{equation}
\label{eq:measurement_model}
\begin{aligned}
\tilde{\mathbf{y}}_{k,i} &:= 
\begin{pmatrix}
    \mathbf{R}^{T}_{i} (\mathbf{b}_{k} - \mathbf{p}_{i}) + \mathbf{n}_{k,i} \\
0 \\
1 \\
\mathbf{0}_{7,1}
\end{pmatrix} \\[1ex]
&= \begin{bmatrix}
\begin{bmatrix}
\mathbf{R}_{i}^{T} & -\mathbf{R}_{i}^{T}\mathbf{v}_i & -\mathbf{R}_{i}^{T} \mathbf{p}_{i} \\
\mathbf{0}_{1,3}  &  1  & 0 \\
\mathbf{0}_{1,3} & 0 & 1
\end{bmatrix} & \mathbf{0}_{5,7} \\
\mathbf{0}_{7,5} & \begin{bmatrix}
\mathbf{I}_{6} & \begin{bmatrix} -\mathbf{b}_{g,i} \\ -\mathbf{b}_{a,i} \end{bmatrix} \\
\mathbf{0}_{1,6} & 1
\end{bmatrix}
\end{bmatrix}  
\begin{pmatrix}
\mathbf{b}_{k} \\
0 \\
1 \\
\mathbf{0}_{7,1}
\end{pmatrix} 
+ \begin{pmatrix} \mathbf{n}_{k,i} \\ \mathbf{0}_{9,1}  \end{pmatrix} \\[1ex]
&:= \mathcal{X}^{-1}_{i}\mathbf{d}_{k} + \tilde{\mathbf{n}}_{k,i},
\end{aligned}
\end{equation}
where $\tilde{\mathbf{n}}_{k,i} \sim \mathcal{N}(\mathbf{0}_{12,1}, \tilde{\mathbf{N}}_{k,i} := \mathrm{diag}(\mathbf{N}_{k,i}, \mathbf{0}_{9,9}))$.

When a set of landmarks $\{\mathbf{b}_k\}_{k=1}^{K}$ is observed simultaneously at the same time instant $t_i$, Eq.~\eqref{eq:so3_meas} can be stacked into a single consolidated measurement vector $\mathbf{y}_i \in \mathbb{R}^{3K}$ given by:
\begin{equation}
\label{eq:so3_meas_stacked}
\mathbf{y}_i := \begin{pmatrix}
    \mathbf{y}_{1,i} \\
    \vdots \\
    \mathbf{y}_{K,i}
\end{pmatrix} = \begin{pmatrix}
\mathbf{R}_i^{T}(\mathbf{b}_1 - \mathbf{p}_i) \\
\vdots \\
\mathbf{R}_i^{T}(\mathbf{b}_K - \mathbf{p}_i)
\end{pmatrix} + \bar{\mathbf{n}}_i,
\end{equation}
where $\bar{\mathbf{n}}_i \sim \mathcal{N}(\mathbf{0}_{3K,1}, \bar{\mathbf{N}}_i := \mathrm{diag}(\mathbf{N}_{1,i}, \ldots, \mathbf{N}_{K,i}))$, while Eq.~\eqref{eq:measurement_model} is stacked to form the augmented measurement vector $\tilde{\mathbf{y}}_i \in \mathbb{R}^{12K}$:
\begin{equation}
    \begin{aligned}
        \tilde{\mathbf{y}}_i &:= \begin{pmatrix} 
    \tilde{\mathbf{y}}_{1,i} \\
    \vdots \\
    \tilde{\mathbf{y}}_{K,i}
\end{pmatrix} = 
\begin{bmatrix}
    \mathcal{X}_i^{-1} & \mathbf{0}_{12,12} & \cdots & \mathbf{0}_{12,12} \\
    \mathbf{0}_{12,12} & \mathcal{X}_i^{-1} & \cdots & \mathbf{0}_{12,12} \\
    \vdots & \vdots & \ddots & \vdots \\
    \mathbf{0}_{12,12} & \mathbf{0}_{12,12} & \cdots & \mathcal{X}_i^{-1} \\
\end{bmatrix} 
\begin{pmatrix} 
    \mathbf{d}_{1} \\
    \vdots \\
    \mathbf{d}_{K}
\end{pmatrix} 
+ \begin{pmatrix}
    \tilde{\mathbf{n}}_{1,i} \\
    \vdots \\
    \tilde{\mathbf{n}}_{K,i}
\end{pmatrix} \\
&:= \mathcal{X}_{\mathrm{aug},i}^{-1} \mathbf{d} + \mathbf{n}_{\mathrm{aug},i}
.\end{aligned}
\end{equation}
In practical applications, the measurement model in Eq.~\eqref{eq:so3_meas} is typically composed with an observation function determined by the sensing modality used to detect the landmarks. For monocular and stereo cameras, this observation function is given by the perspective projection of the landmark onto the image plane, whereas LiDAR and RGB-D sensors provide direct three-dimensional measurements (or equivalently range-based measurements) of the landmark. Although such sensor-specific observation models are omitted here for simplicity, they can be readily incorporated into the proposed estimation framework by composing Eq.~\eqref{eq:so3_meas} with the corresponding observation function~\cite{debeunne2020}.

\section{IEKF and IterIEKF}
\label{sec:invariant}
In this section, we describe the prediction and update steps for both the IEKF and IterIEKF formulations.

\subsection{Prediction Step for Invariant Filtering}
The prediction step for both IEKF and IterIEKF is identical and corresponds to the propagation of the right-invariant error $\boldsymbol{\xi}_i$. This is obtained by linearizing the error kinematics:
\begin{equation}
    \boldsymbol{\xi}_{i|i-1} = \mathrm{Log}_{\mathrm{SE}_2(3)\times \mathbb{R}^{6}}({\mathcal{X}}_{i} \bar{\mathcal{X}}_{i|i-1}^{-1}) \approx 
    \mathbf{A}_{i-1} \boldsymbol{\xi}_{i-1} + \mathbf{B}_{i-1} \mathbf{w}_{i-1}, 
\end{equation}
where $\mathbf{w}_{i-1} \sim \mathcal{N}(\mathbf{0}_{12,1}, \mathbf{Q}_{i-1})$, with the grouped noise covariance defined as $\mathbf{Q}_{i-1} := \mathrm{diag}(\mathbf{Q}_{g,i-1}, \mathbf{Q}_{a,i-1}, \mathbf{Q}_{bg, i - 1}, \mathbf{Q}_{ba, i-1}))$, and the system matrices defined as~\cite{brossard2022,santana2024}:
\begin{equation}
\mathbf{G}_{i-1} = - \bar{\mathbf{R}}_{i|i-1} \mathcal{J}_{r}\Big( (\tilde{\boldsymbol{\omega}}_{\mathcal{I},i-1} - \bar{\mathbf{b}}_{g,i-1})\Delta t \Big) \Delta t,
\end{equation}
\begin{equation}
\mathbf{A}_{i-1} = \begin{bmatrix}
\mathbf{I}_{3} & \mathbf{0}_{3} & \mathbf{0}_{3} & \mathbf{G}_{i-1} & \mathbf{0}_{3} \\
\mathbf{g}^{\wedge}\Delta t & \mathbf{I}_{3} & \mathbf{0}_{3} & \bar{\mathbf{v}}_{i|i-1}^{\wedge}\mathbf{G}_{i-1} & -\bar{\mathbf{R}}_{i-1}\Delta t \\
\mathbf{g}^{\wedge}{\Delta t^{2}}/{2}  & \mathbf{I}_{3}\Delta t & \mathbf{I}_{3} & \bar{\mathbf{p}}_{i|i-1}^{\wedge}\mathbf{G}_{i-1} & -\bar{\mathbf{R}}_{i-1}\Delta t^{2}/2 \\
\mathbf{0}_{3} & \mathbf{0}_{3} & \mathbf{0}_{3} & \mathbf{I}_{3} & \mathbf{0}_{3}  \\
\mathbf{0}_{3} & \mathbf{0}_{3} & \mathbf{0}_{3} & \mathbf{0}_{3} & \mathbf{I}_{3}
\end{bmatrix},
\end{equation}
\begin{equation}
\label{eq:}
\mathbf{B}_{i-1} = 
\begin{bmatrix}
\mathbf{G}_{i-1} & \mathbf{0}_{3} & \mathbf{0}_{3} & \mathbf{0}_{3} \\
\bar{\mathbf{v}}_{i|i-1}^{\wedge}\mathbf{G}_{i-1} & -\bar{\mathbf{R}}_{i-1}\Delta t & \mathbf{0}_{3}  & \mathbf{0}_{3} \\
\bar{\mathbf{p}}_{i|i-1}^{\wedge} \mathbf{G}_{i-1} & -\bar{\mathbf{R}}_{i-1} \Delta t^{2}/2 & \mathbf{0}_{3} & \mathbf{0}_{3} \\
\mathbf{0}_{3}  & \mathbf{0}_{3} & \mathbf{I}_{3}\Delta t & \mathbf{0}_{3} \\
\mathbf{0}_{3}  & \mathbf{0}_{3}  & \mathbf{0}_{3} & \mathbf{I}_{3}\Delta t
\end{bmatrix}.
\end{equation}
From this linearization, the state covariance is propagated via:
\begin{equation}
\label{eq:covariance_propagation}
\mathbf{P}_{i|i-1} \approx \mathbf{A}_{i-1} \mathbf{P}_{i-1} \mathbf{A}_{i-1}^{T} + \mathbf{B}_{i-1} \mathbf{Q}_{i-1} \mathbf{B}_{i-1}^{T} .
\end{equation}
\subsection{Update Step for Invariant Filtering}
We define the innovation for the invariant filtering update step using the algebraic structures from Eq.~\eqref{eq:measurement_model}. For a right-invariant error definition, the innovation vector is expressed as~\cite{barrau2022}:
\begin{equation}
    \label{eq:innov}
    \begin{aligned}
        \mathbf{z}_{k,i} &:= \bar{\mathcal{X}}_{i|i-1} \tilde{\mathbf{y}}_{k,i} - \mathbf{d}_k \\
        &= \bar{\mathcal{X}}_{i|i-1} \mathcal{X}^{-1}_{i} \mathbf{d}_k - \mathbf{d}_k + \bar{\mathcal{X}}_{i|i-1} \tilde{\mathbf{n}}_{k,i} \\
        &= \delta\mathcal{X}_{i}^{-1} \mathbf{d}_k - \mathbf{d}_k + \bar{\mathcal{X}}_{i|i-1} \tilde{\mathbf{n}}_{k,i} \\
        &= \mathrm{Exp}(\boldsymbol{\xi}_{i})^{-1} \mathbf{d}_k - \mathbf{d}_k + \bar{\mathcal{X}}_{i|i-1} \tilde{\mathbf{n}}_{k,i},
    \end{aligned}
\end{equation}
where $\delta\mathcal{X}_{i} := \mathcal{X}_{i} \bar{\mathcal{X}}_{i|i-1}^{-1} := \mathrm{Exp}(\boldsymbol{\xi}_{i})$ represents the right-invariant state error. The standard first-order Lie group approximation $\mathrm{Exp}(\boldsymbol{\xi}_i)^{-1} \approx \mathbf{I} - \boldsymbol{\xi}^{\wedge}_i$ is applied to simplify the deterministic component:
\begin{equation}
    \label{eq:iekf}
    \begin{aligned}
        \mathrm{Exp}(\boldsymbol{\xi}_{i})^{-1}\mathbf{d}_k - \mathbf{d}_k &\approx -\boldsymbol{\xi}_{i}^{\wedge}\mathbf{d}_k \\
        &= -\begin{bmatrix} \boldsymbol{\xi}_{i, \mathrm{SE}_{2}(3)}^{\wedge} & \mathbf{0}_{5,7} \\ \mathbf{0}_{7,5} & \boldsymbol{\xi}_{i,\mathbb{R}^{6}}^{\wedge} \end{bmatrix} \mathbf{d}_{k} \\
        &=   -\begin{bmatrix}
            (\boldsymbol{\xi}_{i}^{\mathbf{R}})^{\wedge} \mathbf{b}_{k} + \boldsymbol{\xi}_{i}^{\mathbf{v}} (0)  + \boldsymbol{\xi}_{i}^{\mathbf{p}}(1) + \boldsymbol{\xi}_{i}^{\mathbf{b}_{g}}(0)  \\
            \mathbf{0}_{1,3}(\mathbf{b}_{k}) + 0(0) + 0(1) \\
            \mathbf{0}_{1,3}(\mathbf{b}_{k}) + 0(0) + 0(1)  \\
			\mathbf{0}_{1,3}(\boldsymbol{\xi}_i^{\mathbf{b}_{g}}) \\
\mathbf{0}_{1,3}(\boldsymbol{\xi}_i^{\mathbf{b}_{a}})
        \end{bmatrix}  \\
        &= \begin{bmatrix}
            \mathbf{b}_{k}^{\wedge}  & \mathbf{0}_{3}  & -\mathbf{I}_{3} & \mathbf{0}_{3,6}\\
            \mathbf{0}_{9,3} & \mathbf{0}_{9,3} & \mathbf{0}_{9,3} & \mathbf{0}_{9,6} \\
        \end{bmatrix}
        \begin{bmatrix}
            \boldsymbol{\xi}_{i}^{\mathbf{R}} \\
            \boldsymbol{\xi}_{i}^{\mathbf{v}} \\
            \boldsymbol{\xi}_{i}^{\mathbf{p}} \\
			\boldsymbol{\xi}_{i}^{\mathbf{b}_{g}} \\
			\boldsymbol{\xi}_{i}^{\mathbf{b}_{a}}
        \end{bmatrix} := \mathbf{H}_k \boldsymbol{\xi}_{i}.
    \end{aligned}
\end{equation}
Substituting this back into Eq.~\eqref{eq:innov} yields the linearized innovation:
\begin{equation}
    \label{eq:linearized_innov}
    \mathbf{z}_{k,i} \approx \mathbf{H}_k \boldsymbol{\xi}_{i} + \hat{\mathbf{n}}_{k,i}, 
\end{equation}
where $\hat{\mathbf{n}}_{k,i} \sim \mathcal{N}(\mathbf{0}_{12,1}, \hat{\mathbf{N}}_{k,i} := \bar{\mathcal{X}}_{i|i-1}\tilde{\mathbf{N}}_{k,i} \bar{\mathcal{X}}_{i|i-1}^{T})$. 

Given that the trailing rows of $\mathbf{z}_{k,i}$ are zero, the formulation can be reduced down to a 3-dimensional subsystem by projecting out the top block:
\begin{equation}
    \tilde{\mathbf{z}}_{k,i} := [\mathbf{z}_{k,i}]_{1:3} \approx \tilde{\mathbf{H}}_{k,i} \boldsymbol{\xi}_{i} + \hat{\tilde{\mathbf{n}}}_{k,i},
\end{equation}
where the reduced measurement matrix is:
\begin{equation}
    \label{eq:tilde_H}
    \tilde{\mathbf{H}}_{k,i} := \begin{bmatrix} \mathbf{b}_{k}^{\wedge} & \mathbf{0}_{3} & -\mathbf{I}_{3} & \mathbf{0}_{3,6} \end{bmatrix},
\end{equation}
and 
$\hat{\tilde{\mathbf{n}}}_{k,i}
\sim \mathcal{N}(
\mathbf{0}_{3,1},
\hat{\tilde{\mathbf{N}}}_{k,i}
:= \bar{\mathbf{R}}_{i|i-1}
\mathbf{N}_{k,i}
\bar{\mathbf{R}}_{i|i-1}^{T}
)$.

For a simultaneous collection of $K$ visual or spatial landmarks, the measurement structures are vertically stacked into a unified system framework:
\begin{equation}
\mathbf{z}_i = 
\begin{pmatrix} \tilde{\mathbf{z}}_{1,i} \\ \vdots \\ \tilde{\mathbf{z}}_{K,i} \end{pmatrix} =
\begin{bmatrix}  
    \mathbf{b}_1^{\wedge} & \mathbf{0}_3 & -\mathbf{I}_3 & \mathbf{0}_{3,6} \\
    \vdots & \vdots & \vdots & \vdots \\
    \mathbf{b}_K^{\wedge} & \mathbf{0}_3 & -\mathbf{I}_3 & \mathbf{0}_{3,6}
\end{bmatrix} 
\boldsymbol{\xi}_{i} + 
\begin{pmatrix} \hat{\tilde{\mathbf{n}}}_{1,i} \\ \vdots \\ \hat{\tilde{\mathbf{n}}}_{K,i} \end{pmatrix} = \mathbf{H}_i \boldsymbol{\xi}_i + \tilde{\mathbf{n}}_i,
\end{equation}
where the collective stacked noise distribution tracks as 
\begin{equation}
    \tilde{\mathbf{n}}_i \sim \mathcal{N}(\mathbf{0}_{3K,1}, \tilde{\mathbf{N}}_i := \mathrm{diag}(\hat{\tilde{\mathbf{N}}}_{1,i}, \ldots, \hat{\tilde{\mathbf{N}}}_{K,i})). 
\end{equation}

The main difference between IterIEKF and IEKF is that instead of performing the linearization in Eq. \eqref{eq:linearized_innov} once per update, 
it applies a Gauss-Newton iterative method to linearize the innovation multiple times until convergence~\cite{goffin2026}. For the right-invariant error, this 
Gauss-Newton sequence was obtained in~\cite{santana2026}. We adapted this loop to be able to handle $K$ visible landmarks, resulting in the following:
\begin{equation}
    \mathbf{\Phi}^{j} = \mathrm{Exp}_{\mathrm{SO}(3)}(-[\boldsymbol{\xi}_{i}^{j}]_{1:3}),
\end{equation}
\begin{equation}
\mathbf{f}_k^{j} = 
[\mathrm{Exp}(-\boldsymbol{\xi}_{i}^{j})\mathbf{d}_k - \mathbf{d}_k]_{1:3}, \, k=1, \ldots, K,
\end{equation}
\begin{equation}
    \mathbf{H}_{i}^{j} = \mathrm{diag}(\underbrace{\mathbf{\Phi}^{j}, \ldots, \mathbf{\Phi}^{j}}_{K \text{ times}}) 
    \mathbf{H}_{i} \mathcal{J}_{r, \mathrm{SE}_2(3) \times \mathbb{R}^{6}}(-\boldsymbol{\xi}_{i}^{j}),
\end{equation}
\begin{equation}
    \mathbf{S}_i^{j} = {\mathbf{H}}_{i}^{j} \mathbf{P}_{i|i-1} ({\mathbf{H}}_{i}^{j})^{T} + \tilde{\mathbf{N}}_{i},
\end{equation}
\begin{equation}
    \mathbf{K}^{j}_{i} = \mathbf{P}_{i|i-1}({\mathbf{H}}_{i}^{j})^{T}
    (\mathbf{S}_i^{j})^{-1},
\end{equation}
\begin{equation}
    \boldsymbol{\xi}_{i}^{j+1} = 
    \mathbf{K}_{i}^{j} ({\mathbf{z}}_{i} - \mathrm{vec}(\mathbf{f}_1^{j}, \ldots, \mathbf{f}_K^{j}) + 
    {\mathbf{H}}_{i}^{j} \boldsymbol{\xi}_{i}^{j}),
\end{equation}
\begin{equation}
  \boldsymbol{\xi}_{i}^{0} := \mathbf{0}_{15,1}.
\end{equation}
Finally, let $l$ denote the last iteration, then the update step for the right-invariant error IterIEKF
is given as follows:
\begin{equation}
    \begin{aligned}
        \hat{\boldsymbol{\xi}}_{i} := \boldsymbol{\xi}_{i}^{l}, \\
    \end{aligned}
\end{equation}
\begin{equation}
    \bar{\mathcal{X}}_{i} := \bar{\mathcal{X}}_{i|i} := \hat{\boldsymbol{\xi}}_{i} \oplus   \bar{\mathcal{X}}_{i|i-1},
\end{equation}
\begin{equation}
    \label{eq:covariance_update}
     \mathbf{P}_{i} := \mathbf{P}_{i|i} = (\mathbf{I}_{15} - \mathbf{K}_{i}^{0} \mathbf{H}_{i}^{0}) \mathbf{P}_{i|i-1},
\end{equation}
To obtain the last iteration, we implemented as a stopping
criterion the condition $\| \boldsymbol{\xi}^{j-1} - \boldsymbol{\xi}^j \| < \delta$, 
where $\delta = 10^{-4}$. We also implemented a second optional stopping criterion based on the 
computation of the loss in \cite[Eq. 34]{santana2026}. If it increases, the algorithm stops.

\section{$\mathrm{SO(3)}$-EKF and Iter$\mathrm{SO(3)}$-EKF}
\label{sec:multiplicative}
In this section, we describe the prediction and update steps for both the standard $\mathrm{SO}(3)$-EKF and the Iter$\mathrm{SO}(3)$-EKF, also widely known as 
Multiplicative-EKF~\cite{markley2003}. These formulations serve as the non-invariant reference baselines for our comparative study.

\subsection{Prediction Step for Multiplicative Filtering}
The prediction step for these filters corresponds to the propagation of the standard multiplicative local error state $\delta \boldsymbol{x}_i$. This is obtained by linearizing the discrete-time error kinematics around the nominal state trajectory~\cite{sola2017}:
\begin{equation}
    \delta \boldsymbol{x}_{i|i-1} = \mathrm{Log}_{\mathrm{SO}(3) \times \mathbb{R}^{12}}({\boldsymbol{x}}_{i} \bar{\boldsymbol{x}}_{i|i-1}^{-1}) \approx 
   \bar{\mathbf{A}}_{i-1} \delta \boldsymbol{x}_{i-1} + \bar{\mathbf{B}}_{i-1} \mathbf{w}_{i-1}, 
\end{equation}
where:
\begin{equation}
\mathbf{F}_{i-1} = -[\bar{\mathbf{R}}_{i-1}( \tilde{\mathbf{a}}_{\mathcal{I},i-1}- \bar{\mathbf{b}}_{a,i-1})]^{\wedge},
\end{equation}
\begin{equation}
\label{eq:mekf_matrices}
\bar{\mathbf{A}}_{i-1} =
\begin{bmatrix}
\mathbf{I}_{3} & \mathbf{0}_{3} & \mathbf{0}_{3} & \mathbf{G}_{i-1} & \mathbf{0}_{3} \\

\mathbf{F}_{i-1}\Delta t & \mathbf{I_{3}} &  \mathbf{0}_{3} & \mathbf{0}_{3} & -\bar{\mathbf{R}}_{i-1}\Delta t \\ 

\mathbf{F}_{i-1} \frac{\Delta t^{2}}{2} & \mathbf{I}_{3}\Delta t & \mathbf{I}_{3} & \mathbf{0}_{3} & -\bar{\mathbf{R}}_{i-1} \frac{\Delta t^{2}}{2} \\
 
\mathbf{0}_{3} & \mathbf{0}_{3} & \mathbf{0}_{3} & \mathbf{I}_{3} & \mathbf{0}_{3} \\

\mathbf{0}_{3}  & \mathbf{0}_{3} & \mathbf{0}_{3} & \mathbf{0}_{3} & \mathbf{I}_{3}
\end{bmatrix},
\end{equation}
\begin{equation}
\bar{\mathbf{B}}_{i-1} = 
\begin{bmatrix}
\mathbf{G}_{i-1} & \mathbf{0}_{3} & \mathbf{0}_{3} & \mathbf{0}_{3} \\
 
\mathbf{0}_{3} & -\bar{\mathbf{R}}_{i-1}\Delta t & \mathbf{0}_{3} & \mathbf{0}_{3} \\

\mathbf{0}_{3} & -\bar{\mathbf{R}}_{i-1} \frac{\Delta t^{2}}{2} & \mathbf{0}_{3} & \mathbf{0}_{3} \\
\mathbf{0}_{3}  & \mathbf{0}_{3} & \mathbf{I}_{3}\Delta t & \mathbf{0}_{3} \\
\mathbf{0}_{3} & \mathbf{0}_{3} & \mathbf{0}_{3} & \mathbf{I}_{3}\Delta t
\end{bmatrix}.
\end{equation}
Using this linearization, the state error covariance $\mathbf{\Sigma}$ is propagated forward in time through:
\begin{equation}
\label{eq:covariance_propagation_mekf}
\mathbf{\Sigma}_{i|i-1} \approx \bar{\mathbf{A}}_{i-1} \mathbf{\Sigma}_{i-1} \bar{\mathbf{A}}_{i-1}^{T} + \bar{\mathbf{B}}_{i-1} \mathbf{Q}_{i-1} \bar{\mathbf{B}}_{i-1}^{T} .
\end{equation}

\subsection{Update Step for Multiplicative Filtering}
For the multiplicative baseline, the measurement innovation is derived directly from the standard observation vector mapping in Eq.~\eqref{eq:so3_meas}. Expanding using standard multiplicative perturbation steps ($\mathbf{R}_i \approx (\delta \boldsymbol{\phi}_i) \oplus \bar{\mathbf{R}}_i$ and $\mathbf{p}_i = \bar{\mathbf{p}}_i + \delta \mathbf{p}_i$), the measurement innovation yields:
\begin{equation}
    \begin{aligned}
\delta \mathbf{z}_{k,i} &:= \mathbf{y}_{k,i} - \bar{\mathbf{R}}_i^{T} (\mathbf{b}_{k} - \bar{\mathbf{p}}_{i}) \\
&= \left(\delta \boldsymbol{\phi}_{i} \oplus \bar{\mathbf{R}}_{i}\right)^{T} (\mathbf{b}_{k} - \bar{\mathbf{p}}_i - \delta \mathbf{p}_i) - \bar{\mathbf{R}}_{i}^{T} (\mathbf{b}_{k} - \bar{\mathbf{p}}_{i}) + \mathbf{n}_{k,i} \\
&\approx \bar{\mathbf{R}}_{i}^{T} (-\delta \boldsymbol{\phi}_{i}^{\wedge})(\mathbf{b}_{k} - \bar{\mathbf{p}}_{i}) - \bar{\mathbf{R}}_{i}^{T} \delta \mathbf{p}_{i} + \mathbf{n}_{k,i} \\
&= \bar{\mathbf{R}}_{i}^{T}(\mathbf{b}_{k} -\bar{\mathbf{p}}_{i})^{\wedge} \delta \boldsymbol{\phi}_{i} - \bar{\mathbf{R}}_{i}^{T} \delta \mathbf{p}_{i} + \mathbf{n}_{k,i},
    \end{aligned}
\end{equation}
which yields the measurement Jacobian matrix:
\begin{equation}
    \label{eq:measurement_jacobian}
    \bar{\mathbf{H}}_{k,i}(\bar{\boldsymbol{x}}_i) :=
    \begin{bmatrix}
        \bar{\mathbf{R}}_i^{T}(\mathbf{b}_k - \bar{\mathbf{p}}_i)^{\wedge} & \mathbf{0}_{3} & -\bar{\mathbf{R}}_i^{T} & \mathbf{0}_{3,6}
    \end{bmatrix}.
\end{equation}
For a tracking sequence containing $K$ visible spatial features, the observation parameters are vertically concatenated:
\begin{equation}
    \begin{aligned}
        \mathbf{f}_k(\bar{\boldsymbol{x}}_i) &= \bar{\mathbf{R}}_i^{T} (\mathbf{b}_k - \bar{\mathbf{p}}_i), \quad k = 1, \ldots, K, \\
        \delta \mathbf{z}_i &= \mathbf{y}_i - \begin{bmatrix} \mathbf{f}_1(\bar{\boldsymbol{x}}_i) \\ \vdots \\ \mathbf{f}_K(\bar{\boldsymbol{x}}_i) \end{bmatrix} \approx \bar{\mathbf{H}}_i(\bar{\boldsymbol{x}}_i)\delta \boldsymbol{x}_i + \bar{\mathbf{n}}_i,
    \end{aligned}
\end{equation}
where the complete augmented measurement Jacobian block $\bar{\mathbf{H}}_{i}(\bar{\boldsymbol{x}}_i) \in \mathbb{R}^{3K \times 15}$ is structured as:
\begin{equation}
    \bar{\mathbf{H}}_{i}(\bar{\boldsymbol{x}}_i) :=
    \begin{bmatrix}
        \bar{\mathbf{R}}_i^{T}(\mathbf{b}_1 - \bar{\mathbf{p}}_i)^{\wedge} & \mathbf{0}_{3} & -\bar{\mathbf{R}}_i^{T} & \mathbf{0}_{3,6} \\
        \vdots & \vdots & \vdots & \vdots \\
        \bar{\mathbf{R}}_i^{T}(\mathbf{b}_K - \bar{\mathbf{p}}_i)^{\wedge} & \mathbf{0}_{3} & -\bar{\mathbf{R}}_i^{T} & \mathbf{0}_{3,6}
    \end{bmatrix}.
\end{equation}
Importantly, as highlighted by contrasting Eq.~\eqref{eq:measurement_jacobian} with the invariant formulation in Eq.~\eqref{eq:tilde_H}, the multiplicative measurement matrix explicitly depends on the absolute state estimate trajectories ($\bar{\mathbf{R}}_i, \bar{\mathbf{p}}_i$).

The update sequence for the Iter$\mathrm{SO}(3)$-EKF tracks a localized Gauss-Newton optimization iteration scheme. We adapt the single-measurement sequence from~\cite[Eq. 59]{santana2026} into a stacked multi-landmark framework evaluated at optimization index $j$:
\begin{equation}
\mathbf{\Gamma}^{j}_{i} = \left(\delta \boldsymbol{\phi}_i^{j} \oplus \bar{\mathbf{R}}_{i|i-1}\right)^{T},
\end{equation}
\begin{equation}
\begin{aligned}
        \bar{\mathbf{H}}_{i}^{j} =
        \begin{bmatrix}
            \mathbf{\Gamma}^{j}_{i}({\mathbf{b}}_{1} - \bar{\mathbf{p}}_{i|i-1} -  \delta \mathbf{p}_i^{j})^{\wedge} \mathcal{J}_{r, \mathrm{SO}(3)}(-\delta \boldsymbol{\phi}_i^{j}) & \mathbf{0}_{3} & -\mathbf{\Gamma}_i^{j} & \mathbf{0}_{3,6} 
             \\
           \vdots & \vdots & \vdots & \vdots  \\
            \mathbf{\Gamma}^{j}_{i}({\mathbf{b}}_{K} - \bar{\mathbf{p}}_{i|i-1} -  \delta \mathbf{p}_i^{j})^{\wedge} \mathcal{J}_{r, \mathrm{SO}(3)}(-\delta \boldsymbol{\phi}_i^{j}) & \mathbf{0}_{3} & -\mathbf{\Gamma}_i^{j} & \mathbf{0}_{3,6}
        \end{bmatrix},
\end{aligned}
\end{equation}
\begin{equation}
    \bar{\mathbf{S}}_i^{j} = \bar{\mathbf{H}}_{i}^{j} \mathbf{\Sigma}_{i|i-1} (\bar{\mathbf{H}}_{i}^{j})^{T} + \bar{\mathbf{N}}_i,
\end{equation}
\begin{equation}
    \bar{\mathbf{K}}^{j}_{i} = \mathbf{\Sigma}_{i|i-1}(\bar{\mathbf{H}}_{i}^{j})^{T} (\bar{\mathbf{S}}_i^{j})^{-1},
\end{equation}
\begin{equation}
    \delta \boldsymbol{x}_{i}^{j+1} = \bar{\mathbf{K}}_{i}^{j} \left( {\mathbf{y}}_{i} - \begin{bmatrix} \mathbf{f}_1(\bar{\boldsymbol{x}}_i^j) \\ \vdots \\ \mathbf{f}_K(\bar{\boldsymbol{x}}_i^j) \end{bmatrix} + \bar{\mathbf{H}}_{i}^{j} \delta \boldsymbol{x}_{i}^{j} \right),
\end{equation}
initialized at the start of the optimization sequence with $\delta \boldsymbol{x}_{i}^{0} := \mathbf{0}_{15,1}$. Letting $l$ denote the final iteration step satisfying convergence, the posteriors are computed as:
\begin{equation}
  \bar{\boldsymbol{x}}_{i} :=  \bar{\boldsymbol{x}}_{i|i} = \delta \boldsymbol{x}_{i}^{l} \oplus \bar{\boldsymbol{x}}_{i|i-1},
\end{equation}
\begin{equation}
   \mathbf{\Sigma}_{i} := \mathbf{\Sigma}_{i|i} = (\mathbf{I}_{15} - \bar{\mathbf{K}}_{i}^{l} \bar{\mathbf{H}}_{i}^{l}) \mathbf{\Sigma}_{i|i-1}.
\end{equation}
The execution loop for the Iter$\mathrm{SO}(3)$-EKF employs identical convergence stopping criteria thresholds as those configured for the IterIEKF.
\section{Results}
\label{sec:results}
\begin{figure}[h!]
    \centering
    \includegraphics[scale=0.9]{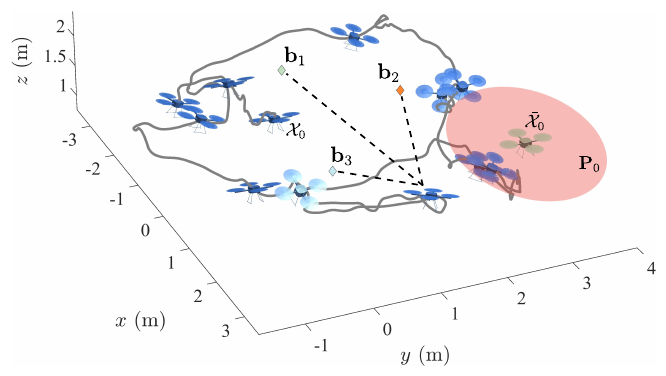}
    \caption{\texttt{V2\_01\_easy} dataset trajectory (gray), MAV true state (blue), landmarks (diamonds), and a single MAV sample of the corrupted initial state configuration for the invariant estimators (green).}
    \label{fig:scene}
\end{figure}

We evaluate the performance of the proposed IterIEKF using a structured Monte Carlo simulation framework. The ground-truth inertial measurements are generated by solving the inverse kinematic problem over a known reference trajectory. Specifically, we extract the true state trajectories from the \texttt{V2\_01\_easy} dataset\footnote{\url{https://github.com/rpng/open_vins/blob/master/ov_data/euroc_mav/V2_01_easy.csv}} of the EuRoC MAV repository~\cite{burri2016} and reconstruct ideal IMU inputs that satisfy the discrete-time system kinematics:
\begin{equation}
\label{eq:imu_reconstruction}
\begin{aligned}
\mathcal{X}_{i+1} &= \mathcal{W}_i \Phi({\mathcal{X}}_{i}) \mathcal{Y}_i(\boldsymbol{\omega}_{\mathcal{I},i}, \mathbf{a}_{\mathcal{I},i}) \\  
\Rightarrow \left( \mathcal{W}_i \Phi({\mathcal{X}}_i) \right)^{-1} \mathcal{X}_{i+1} &= \mathcal{Y}_i(\boldsymbol{\omega}_{\mathcal{I},i}, \mathbf{a}_{\mathcal{I},i}) \\  
\Rightarrow \quad \left( \mathrm{Exp}_{\mathrm{SO}(3)}(\mathbf{a}),\, \mathbf{b} \right) &= \left( \mathrm{Exp}_{\mathrm{SO}(3)}(\boldsymbol{\omega}_{\mathcal{I},i}\Delta t),\, \mathbf{L}\mathbf{a}_{\mathcal{I},i} \right), \\
\Rightarrow \quad \boldsymbol{\omega}_{\mathcal{I},i} = \frac{\mathbf{a}}{\Delta t}, \quad &\mathbf{a}_{\mathcal{I},i} = \mathbf{L}^{\dagger} \mathbf{b}, \quad \mathcal{W}_i, \Phi, \mathcal{Y}_i \in \mathrm{SE}_2(3), 
\end{aligned}
\end{equation}
where the kinematic prediction model from Eq.~\eqref{eq:prediction_step} is reformulated as a composition of algebraic elements on $\mathrm{SE}_2(3)$~\cite{brossard2022}. The coupling matrix $\mathbf{L}$ is defined as $\mathbf{L} = [\mathbf{I}_3 \Delta t; \mathbf{I}_3 \frac{\Delta t^2}{2}]$, and $\mathbf{L}^{\dagger} = (\mathbf{L}^{T}\mathbf{L})^{-1}\mathbf{L}^{T}$ represents its left pseudo-inverse. In addition to obtaining the true inertial measurements, 
we also artificially placed three landmarks in the world frame, $\mathbf{b}_1 = [-2.0,1.0,1.6]^{T}$, $\mathbf{b}_2 = [0.0, 2.0, 2.0]^{T}$ and $\mathbf{b}_3 = [1.0, 0.5, 1.5]^{T}$.

Across $N=50$ localized Monte Carlo realizations, we corrupt the true initial state configuration, the proprioceptive inertial sequences, and the exteroceptive landmark updates with zero-mean white Gaussian noise structured as:
\begin{equation}
 \begin{aligned}
 \mathbf{P}_{0} = \text{diag}\left(\left(\frac{\pi}{4}\right)^2 \mathbf{I}_3, \, 1.0^{2} \mathbf{I}_3, \, 2.0^{2} \mathbf{I}_3, \, 10^{-6}\mathbf{I}_3, 10^{-6} \mathbf{I}_3 \right), \\
 \mathbf{w}_{a,i} \sim \mathcal{N}(\mathbf{0}_{3,1}, 1.6 \times 10^{-3}\mathbf{I}_3 ),\, 
 \mathbf{w}_{g,i} \sim \mathcal{N}(\mathbf{0}_{3,1}, 4.0 \times 10^{-6} \mathbf{I}_3), \\
 \mathbf{w}_{bg,i} , \mathbf{w}_{ba,i} \sim \mathcal{N}(\mathbf{0}_{3,1}, 1.0 \times 10^{-6} \mathbf{I}_3),\, 
 \mathbf{n}_{k,i} \sim \mathcal{N}(\mathbf{0}_{3,1}, 1.0 \times 10^{-3} \mathbf{I}_3).
 \end{aligned}
\end{equation}
During each initialization pass, the true baseline state is perturbed by an error displacement sample $\boldsymbol{\xi}_0 \sim \mathcal{N}(\mathbf{0}_{15,1}, \mathbf{P}_0)$. This generates the nominal starting state for the invariant filters via $\bar{\mathcal{X}}_0 = (-\boldsymbol{\xi}_0) \oplus \mathcal{X}_0$, while the multiplicative variants initialize through $\bar{\boldsymbol{x}}_0 = (-\delta \boldsymbol{\boldsymbol{x}}_0) \oplus \boldsymbol{x}_0$, with $\delta \boldsymbol{x}_0 = \mathbf{J} \boldsymbol{\xi}_0$ mapped via the state-space transformation Jacobian defined in~\cite[Eq. 76]{santana2026}. The trajectory trace and a sample visualization of this initialization are shown in Fig.~\ref{fig:scene}. The simulated IMU feeds operate at $200\,\mathrm{Hz}$, whereas landmark observations update at a reduced rate of $1\,\mathrm{Hz}$ with all landmarks visible during the entire trajectory, i.e., $K = 3$.

Filter statistical consistency is evaluated over all time horizons using the Average Normalized Estimation Error Squared (NEES) metric:
\begin{equation}
    \label{eq:nees}
    \begin{aligned}
        \mathrm{NEES}_i^{\mathrm{SE_2}(3) \times \mathbb{R}^{6}} &= \frac{1}{N} \sum_{n=1}^{N} (\boldsymbol{\xi}_{i}^{n})^{T} (\mathbf{P}_{i}^{n})^{-1} \boldsymbol{\xi}_{i}^{n}, \\
        \mathrm{NEES}_i^{\mathrm{SO}(3) \times \mathbb{R}^{12}} &= \frac{1}{N} \sum_{n=1}^{N} ({\delta \boldsymbol{x}}_{i}^{n})^{T} (\mathbf{\Sigma}_{i}^{n})^{-1} {\delta  \boldsymbol{x}}_{i}^{n}.
    \end{aligned}
\end{equation}
Trajectory accuracy is quantified through the Mean Absolute Error (MAE) evaluated across the system's observable state space, including the body-frame linear velocity ($\bar{\mathbf{R}}_{i|i}^{T} \bar{\mathbf{v}}_{i|i}$), global gravity vector orientation ($\bar{\mathbf{u}}_{i|i}$), and Euler attitude parameters (yaw $\bar{\psi}_{i|i}$, pitch $\bar{\phi}_{i|i}$, and roll $\bar{\theta}_{i|i}$). The error metric tracks as:
\begin{equation}
    \mathrm{MAE}_{\mathbf{x}} = \frac{1}{M} \sum_{i=1}^{M} e(\bar{\mathbf{x}}_{i|i}, \mathbf{x}_{i}),
\end{equation}
where $M$ defines the total number of timestamps across the mission duration. For position, velocity, and attitude states, $e(\cdot)$ corresponds to the standard Euclidean distance metric $e := \| \mathbf{x}_i - \bar{\mathbf{x}}_{i|i} \|$. For tracking the gravity vector direction, the metric is defined as the angular displacement $e := \arccos(\bar{\mathbf{u}}_{i|i} \cdot \mathbf{u}_i)$ to provide a physically intuitive representation of angular drift.

\subsection{Accuracy and Consistency of the Observable State}
\begin{figure}[h!]
		\centering
        \includegraphics[scale=0.320]{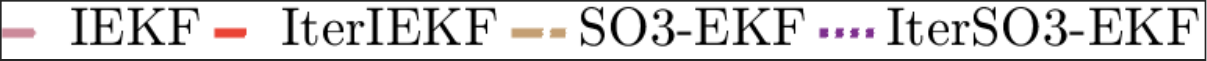}
        \subfloat[\protect\label{fig:position}]{
            \includegraphics[width=0.45\textwidth]{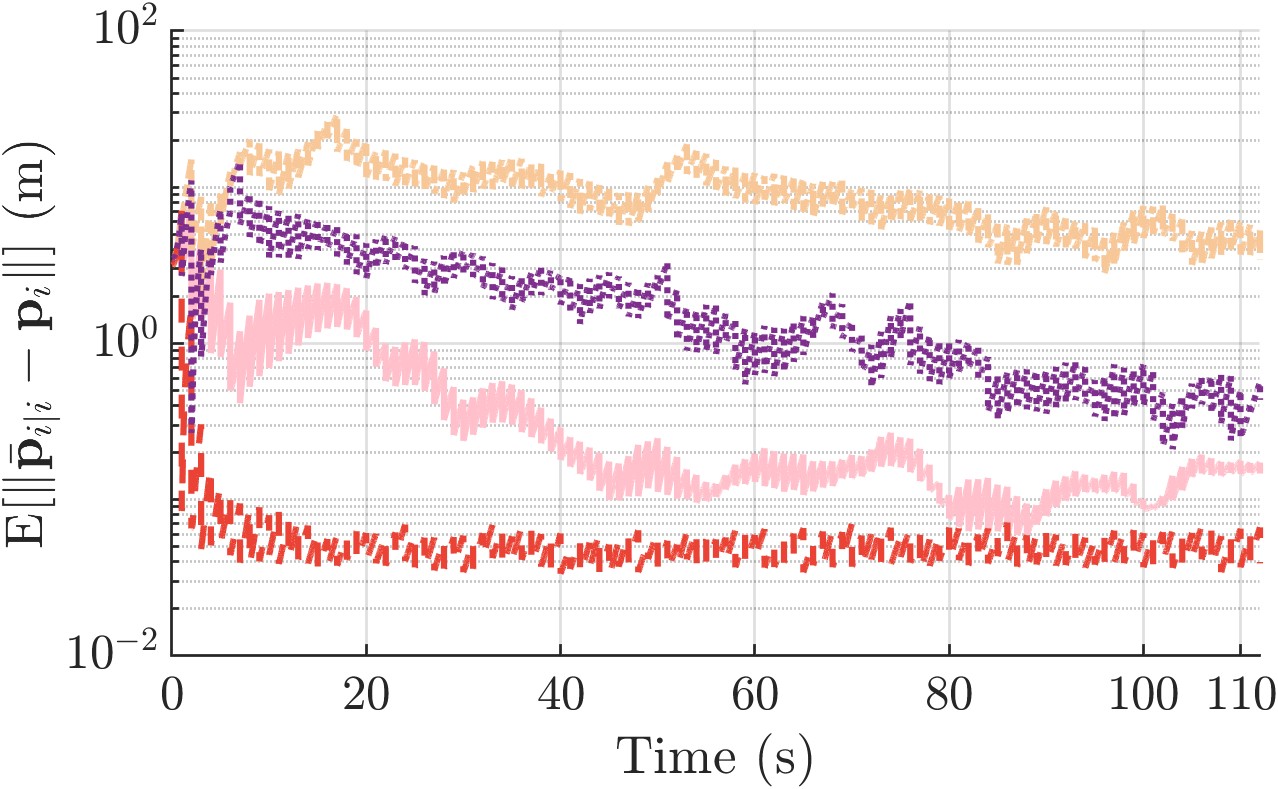}
    }
    \subfloat[\protect\label{fig:velocity}]{
        \includegraphics[width=0.45\textwidth]{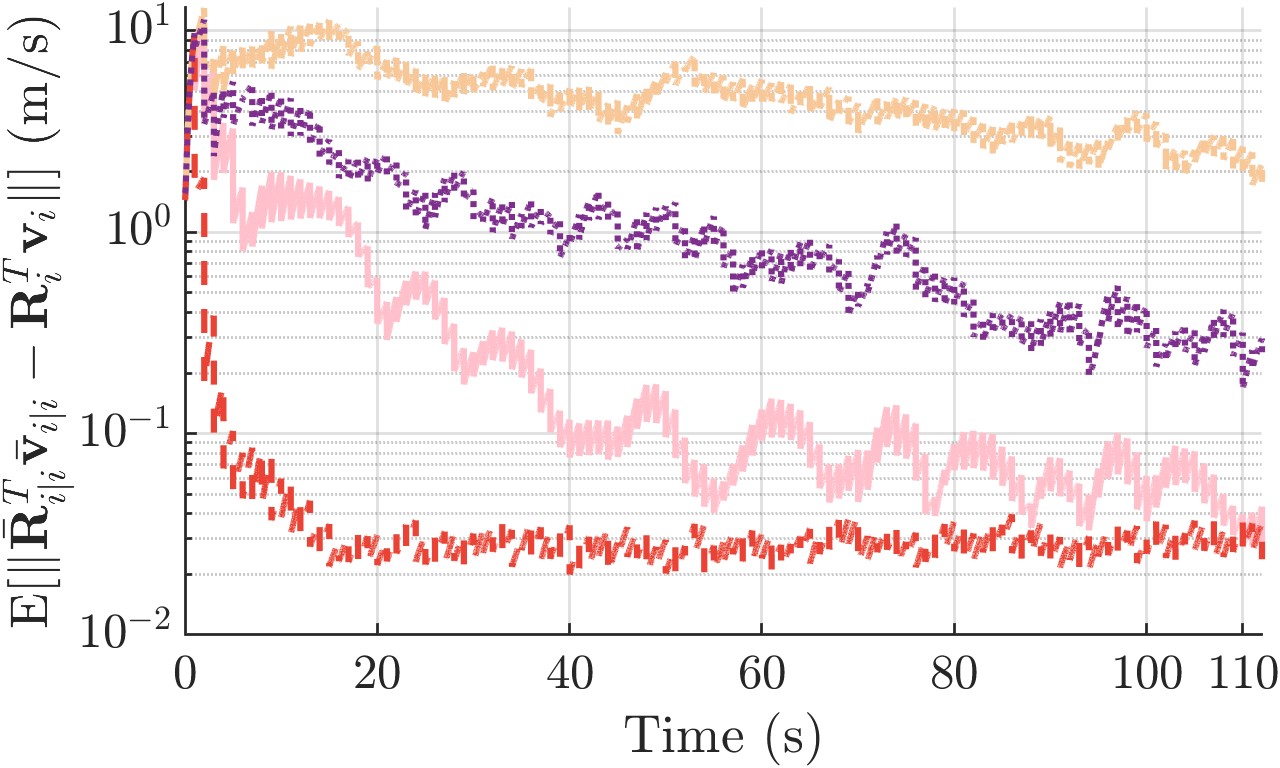}
    }
   \hfill
    \subfloat[\protect\label{fig:gravity_direction}]{
        \includegraphics[width=0.45\textwidth]{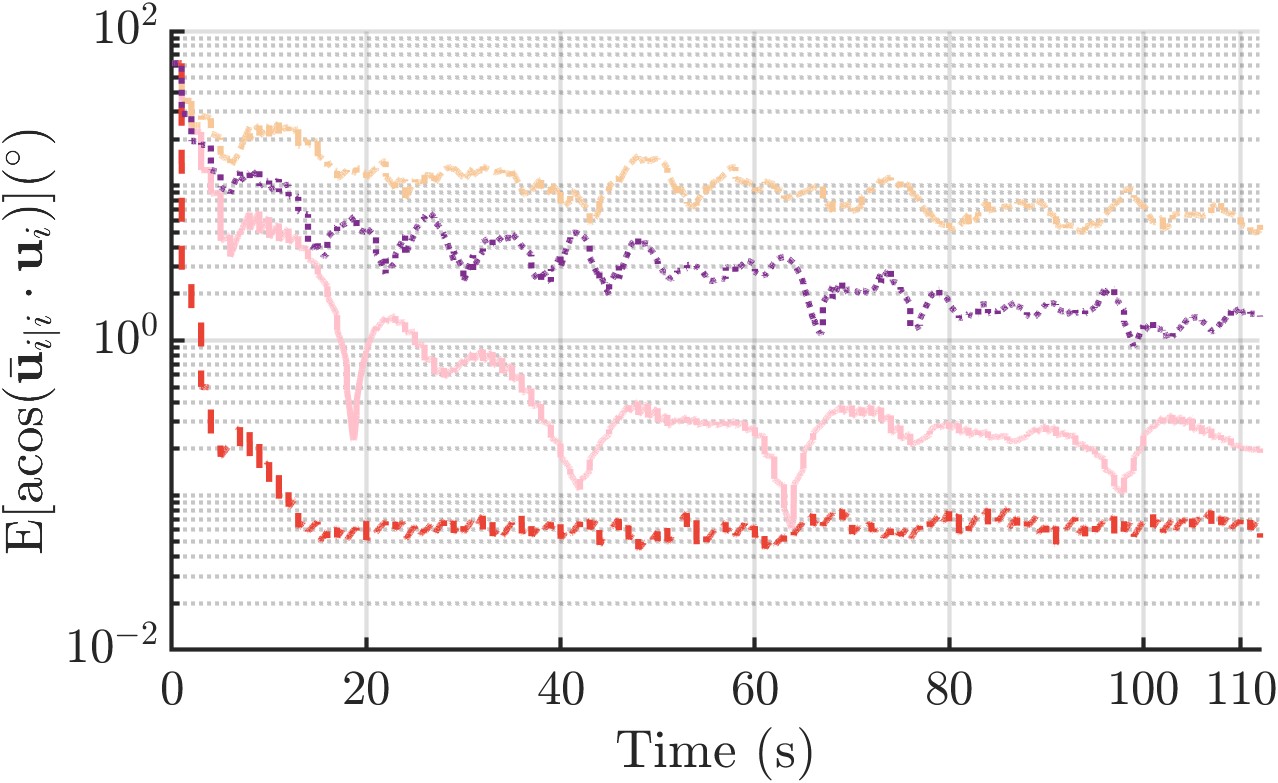}
    }
    \subfloat[\protect\label{fig:nees}]{
        \includegraphics[width=0.45\textwidth]{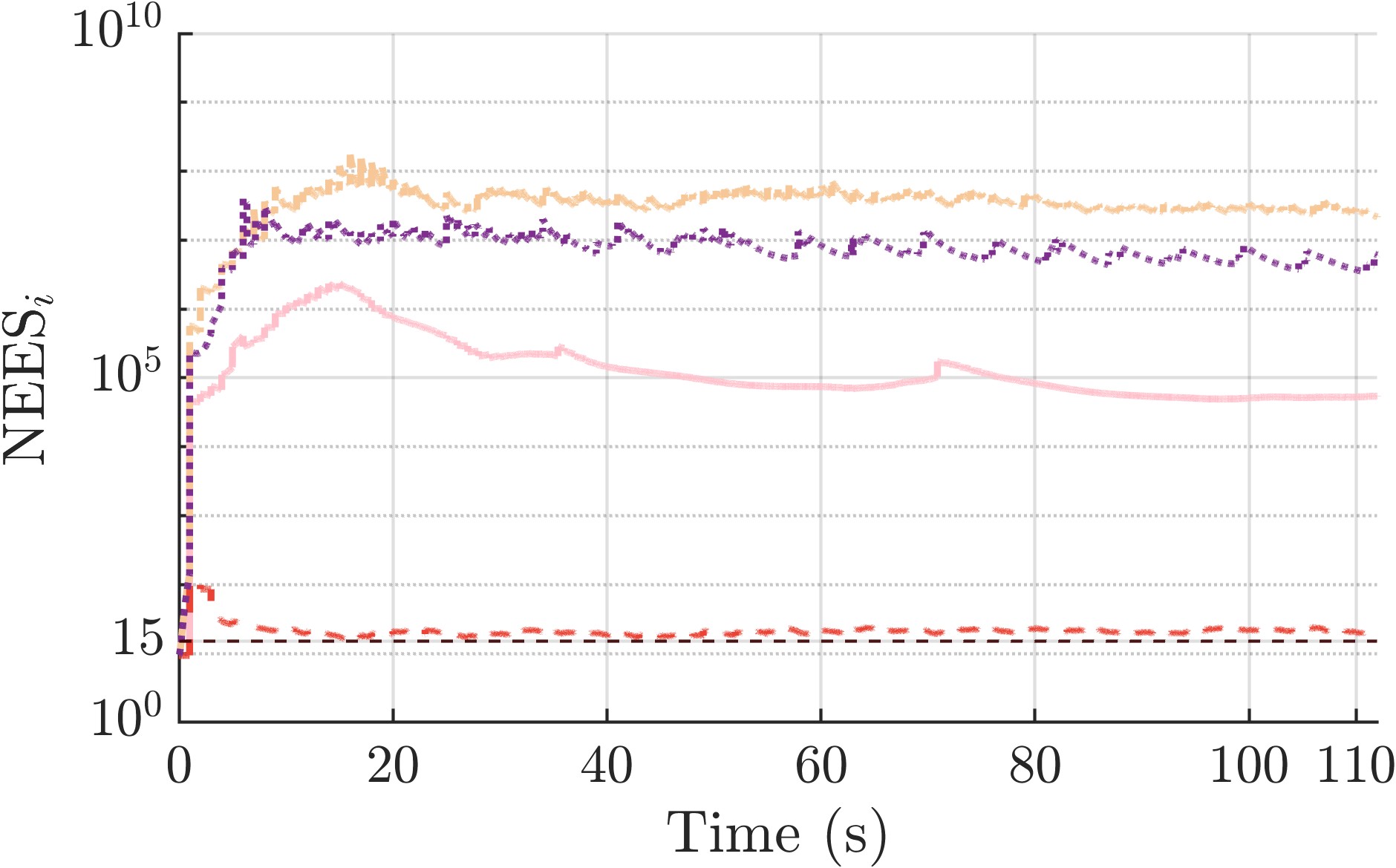}
    }
    \caption{Results of the numerical Monte Carlo experiment on the \texttt{V2\_01\_easy} dataset. 
        a) Average error for the position. 
        b) Average error for the velocity in the base frame.
        c) Average error for the gravity direction.
        d) Average NEES.}
    \label{fig:result_scenario_i}
\end{figure}

The quantitative performance metrics across the $N=50$ Monte Carlo simulation trials, consolidated in Table~\ref{tab:mae_results}, reveal a performance disparity between the evaluated filtering paradigms under initial state corruption. The classical $\mathrm{SO}(3)$-EKF exhibits the poorest performance, accumulating a substantial MAE of $9.363\,\mathrm{m}$ in position and $4.918\,\mathrm{m/s}$ in linear velocity due to the well-documented false observability problem, where state-dependent measurement Jacobians erroneously introduce spurious information along unobservable directions. While the $\text{IterSO}(3)$-EKF mitigates local linearization errors through iterative refinement loops, reducing the position and velocity MAE to $2.1\,\mathrm{m}$ and $1.331\,\mathrm{m/s}$, respectively, it lacks geometric error parameterization and remains significantly less accurate than the standard single-step $\mathrm{IEKF}$ ($0.511\,\mathrm{m}$). In contrast, the proposed $\text{IterIEKF}$ achieves state-of-the-art performance, outperforming the baseline $\mathrm{IEKF}$ by reducing position error to $0.096\,\mathrm{m}$ (an $81.12\%$ reduction) and velocity error to $0.095\,\mathrm{m/s}$ (an $80.14\%$ reduction). This remarkable tracking fidelity highlights the benefit of fusing Lie-group error dynamics on $\mathrm{SE}_2(3)$, which structurally decouple the error kinematics from the absolute trajectory to guarantee correct observability properties, with localized Gauss-Newton optimization loops that drive the estimated state precisely to the true intersection of the environmental landmark manifolds.
The convergence of the estimated position, velocity and gravity direction are shown in Figs.~\ref{fig:position}, \ref{fig:velocity} and \ref{fig:gravity_direction}. The $\text{IterIEKF}$ demonstrates superior convergence characteristics, rapidly aligning with the ground truth trajectory within the first $10\,\mathrm{s}$ of operation, while the $\mathrm{IEKF}$ exhibits a slower convergence rate and larger steady-state error. The $\text{IterSO}(3)$-EKF, although improving upon the $\mathrm{SO}(3)$-EKF, still fails to achieve comparable accuracy due to its inherent limitations in handling non-linearities and observability issues.

In Fig.~\ref{fig:nees} we show the NEES metric for the different filters, where the $\text{IterIEKF}$ maintains a NEES value close to $15$ (the error-state dimension), indicating good consistency, while the $\mathrm{IEKF}$ and $\text{IterSO}(3)$-EKF show higher NEES values, suggesting overconfidence in their estimates. The $\mathrm{SO}(3)$-EKF exhibits the highest NEES values, confirming its poor consistency and reliability in state estimation under severe initial state corruption.
\begin{table}
\caption{Mean Absolute Error (MAE) and relative errors to IEKF (\%) for different filtering methods. Smallest MAE is marked in bold.}\label{tab1}
\label{tab:mae_results}
\centering
\begin{tabular}{|l|c|c|c|}
\hline
Filters & $\mathrm{MAE}_{\mathrm{velocity}}$ (m/s) & $\mathrm{MAE}_{\mathrm{gravity \, direction}}$ ($^{\circ}$) & $\mathrm{MAE}_{\mathrm{position}}$ (m) \\
\hline
IEKF & 
  \begin{tabular}{c} 0.48 \\ \small (0\%) \end{tabular} & 
  \begin{tabular}{c} 2.042 \\ \small (0\%) \end{tabular} & 
  \begin{tabular}{c} 0.511 \\ \small (0\%) \end{tabular} \\
\hline
IterIEKF & 
  \begin{tabular}{c} \textbf{0.095} \\ \small (-80.135\%) \end{tabular} & 
  \begin{tabular}{c} \textbf{0.661} \\ \small (-67.641\%) \end{tabular} & 
  \begin{tabular}{c} \textbf{0.096} \\ \small (-81.117\%) \end{tabular} \\
\hline
$\mathrm{SO}(3)$-EKF & 
  \begin{tabular}{c} 4.918 \\ \small (+925.35\%) \end{tabular} & 
  \begin{tabular}{c} 11.244 \\ \small (+450.7\%) \end{tabular} & 
  \begin{tabular}{c} 9.363 \\ \small (+1732.8\%) \end{tabular} \\
\hline
  Iter$\mathrm{SO(3)}$-EKF & 
  \begin{tabular}{c} 1.331 \\ \small (+177.49\%) \end{tabular} & 
  \begin{tabular}{c} 4.439 \\ \small (+117.41\%) \end{tabular} & 
  \begin{tabular}{c} 2.1 \\ \small (+311.15\%) \end{tabular} \\
\hline
\end{tabular}
\end{table}

\subsection{Attitude Estimation}
\begin{figure}[h!]
		\centering
        \includegraphics[scale=0.320]{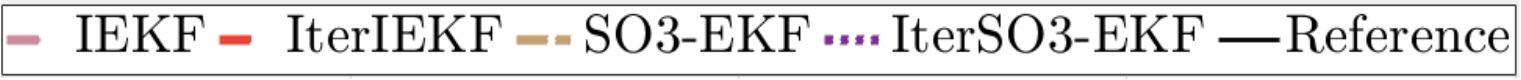}
        \subfloat[\protect\label{fig:roll}]{
            \includegraphics[width=0.32\textwidth]{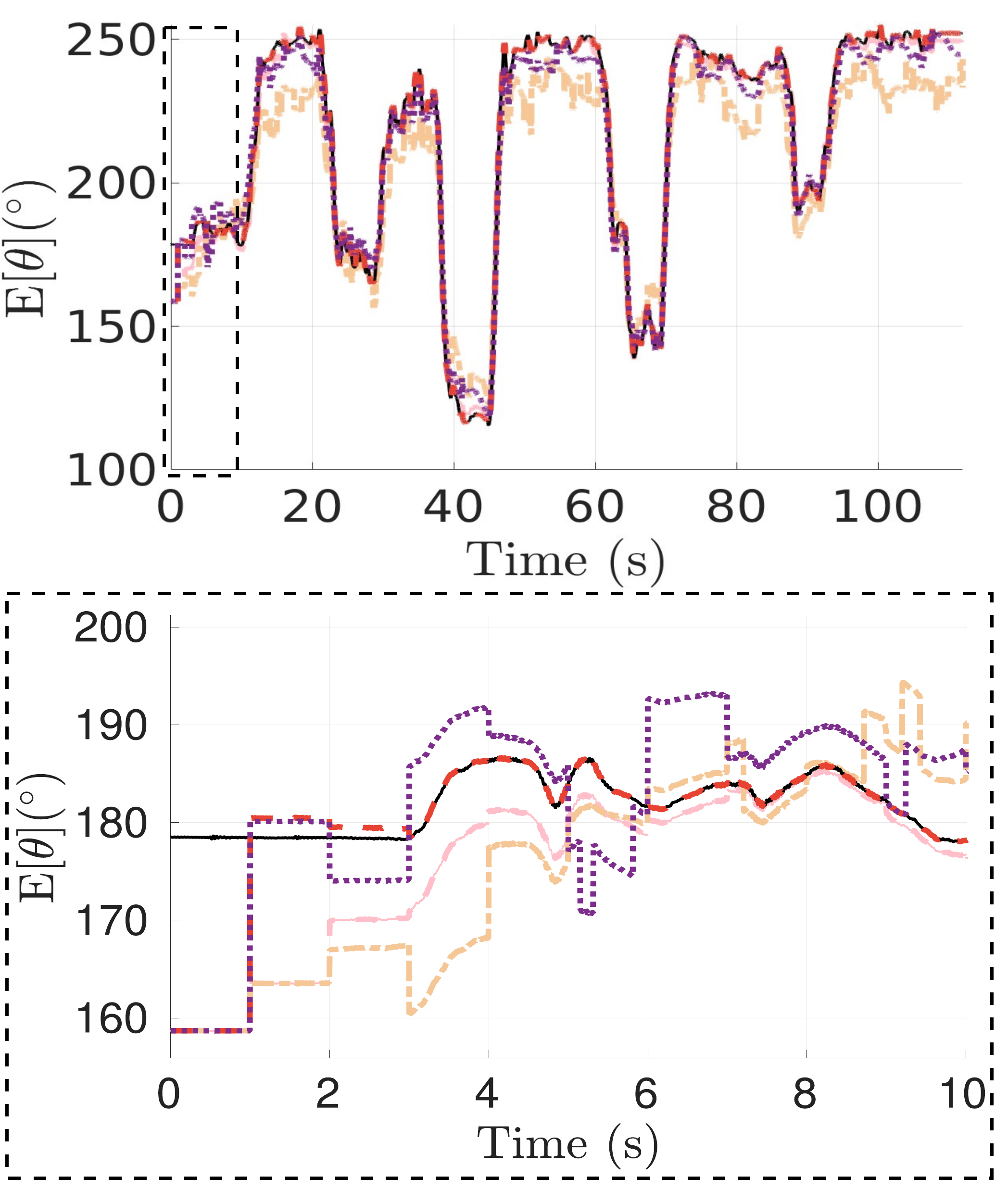}
    }
    \subfloat[\protect\label{fig:pitch}]{
        \includegraphics[width=0.32\textwidth]{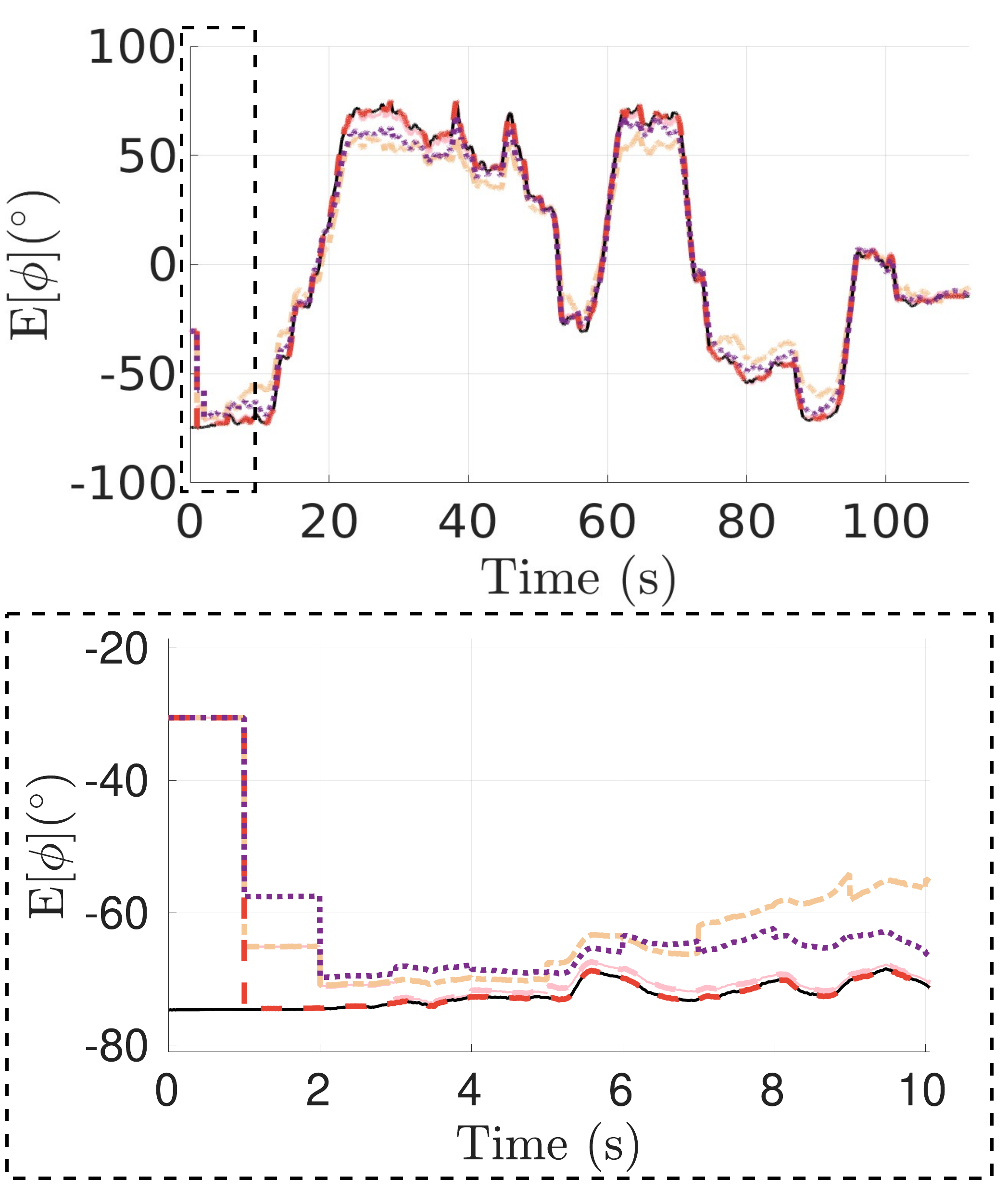}
    }
    \subfloat[\protect\label{fig:yaw}]{
        \includegraphics[width=0.32\textwidth]{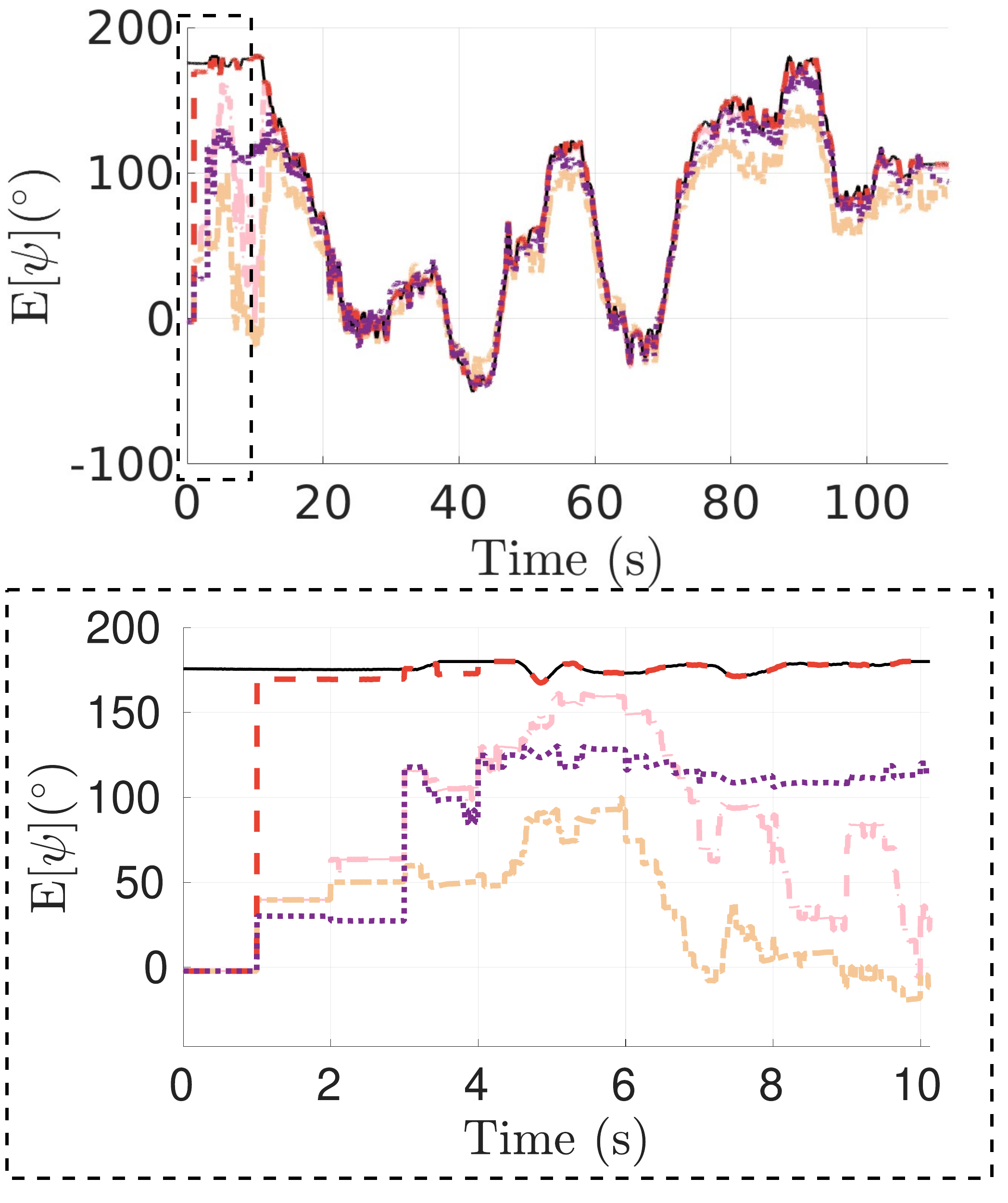}
    }
    \caption{
        Average attitude estimates obtained with different filters across all realizations. 
            a) Estimate of the roll angle. b) Estimate of the pitch angle. c) Estimate of the yaw angle.}
    \label{fig:attitude_estimation}
\end{figure}
The average attitude estimation (roll, pitch and yaw angles) computed in all Monte Carlo realizations is shown in Fig.~\ref{fig:attitude_estimation}. Overall, the proposed $\text{IterIEKF}$ achieves the fastest convergence rate, settling almost $10\,\text{seconds}$ ahead of the standard, single-step $\mathrm{IEKF}$ (see Fig.~\ref{fig:yaw}). Interestingly, the $\text{IterIEKF}$ exhibits a slightly prolonged settling time when converging the roll angle (Fig.~\ref{fig:roll}), whereas it experiences highly accelerated convergence characteristics in the pitch channel (Fig.~\ref{fig:pitch}). 
Furthermore, both the classical $\mathrm{SO}(3)$-EKF and its iterated counterpart, the $\text{IterSO}(3)$-EKF, demonstrate significantly slower response dynamics, though the $\text{IterSO}(3)$-EKF noticeably outperforms the baseline EKF due to its local optimization loops. These tracking profiles confirm that the $\text{IterIEKF}$ possesses superior robustness against severe initial state corruption, providing the most reliable and highly consistent attitude estimation performance among all evaluated filters.
\subsection{Bias Convergence Analysis}
\begin{figure}[h!]
		\centering
        \includegraphics[scale=0.375]{./pictures/main_result/legend_wo_reference.pdf}
        \subfloat[\protect\label{fig:gyro_bias_x}]{
            \includegraphics[width=0.48\textwidth]{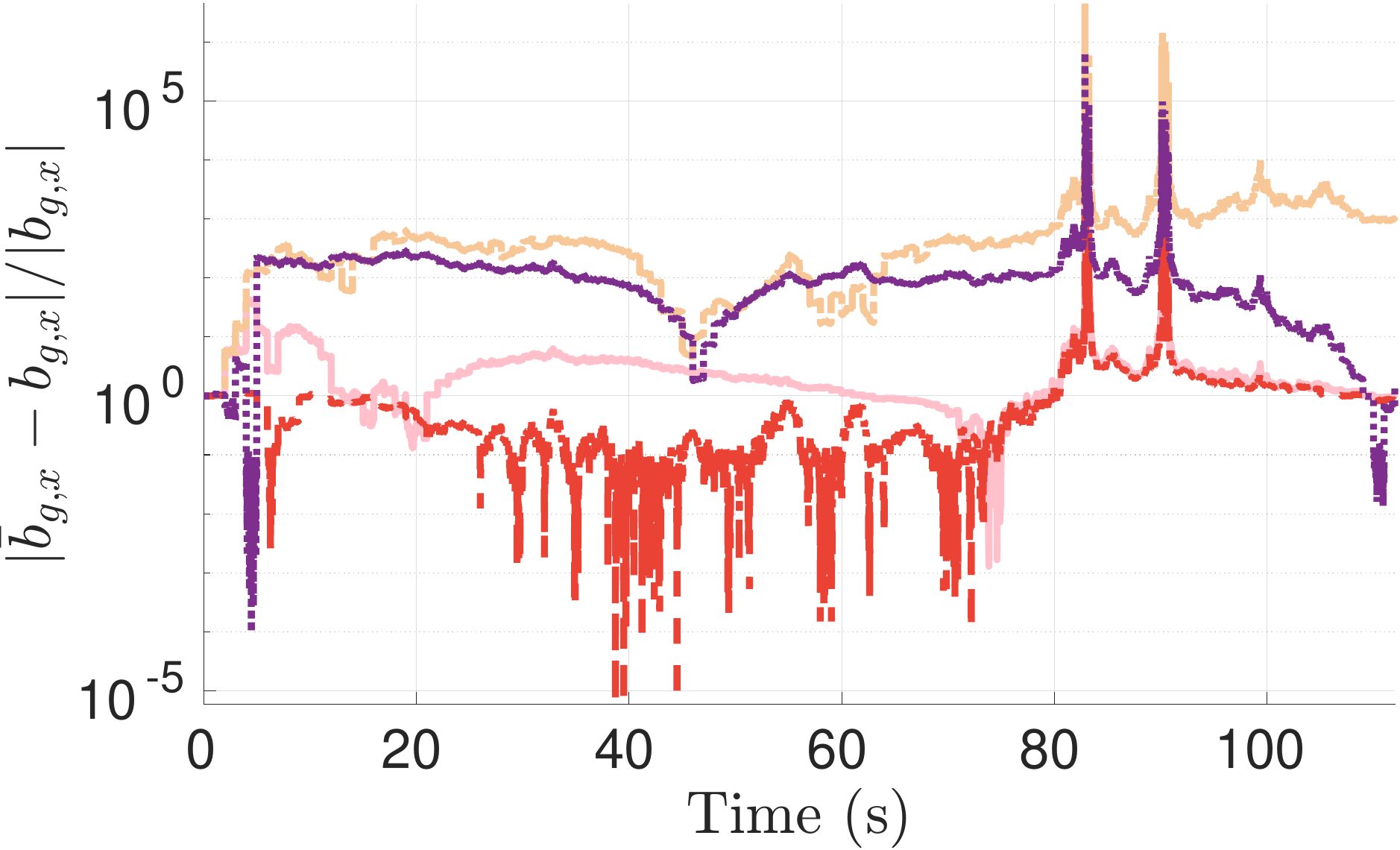}
    }
    \subfloat[\protect\label{fig:gyro_bias_y}]{
        \includegraphics[width=0.48\textwidth]{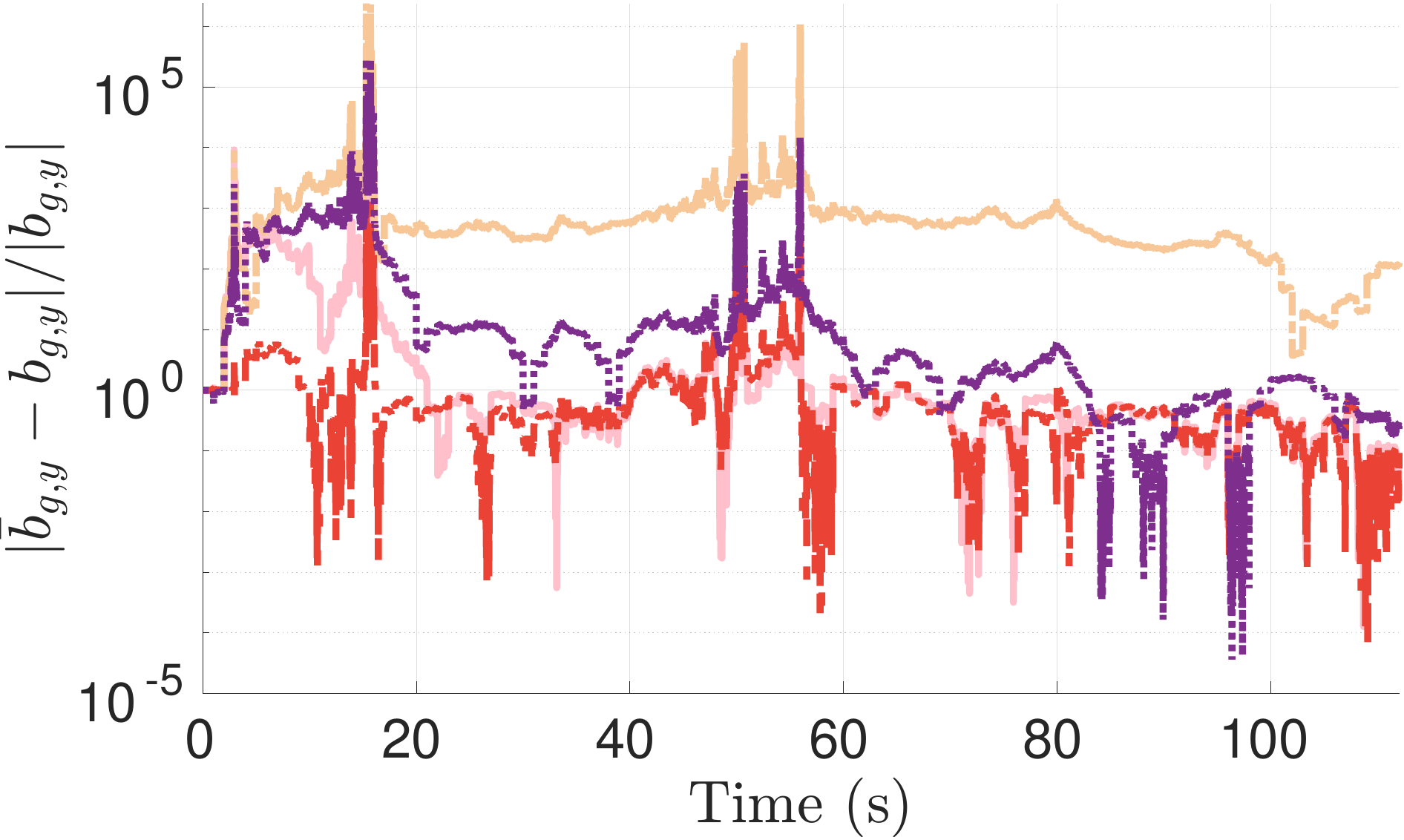}
    }
   \hfill
    \subfloat[\protect\label{fig:gyro_bias_z}]{
        \includegraphics[width=0.48\textwidth]{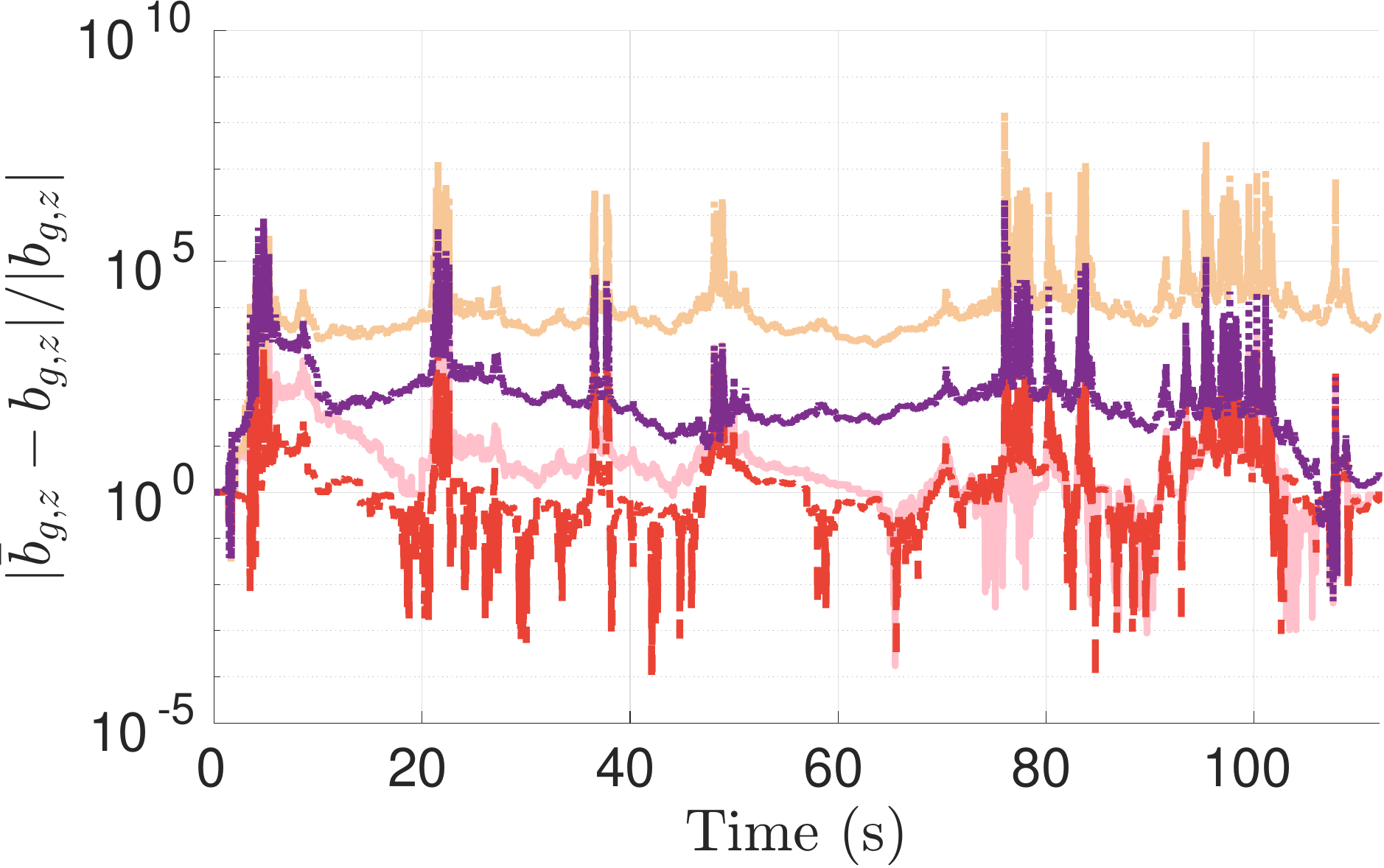}
    }
    \subfloat[\protect\label{fig:accel_bias_x}]{
        \includegraphics[width=0.48\textwidth]{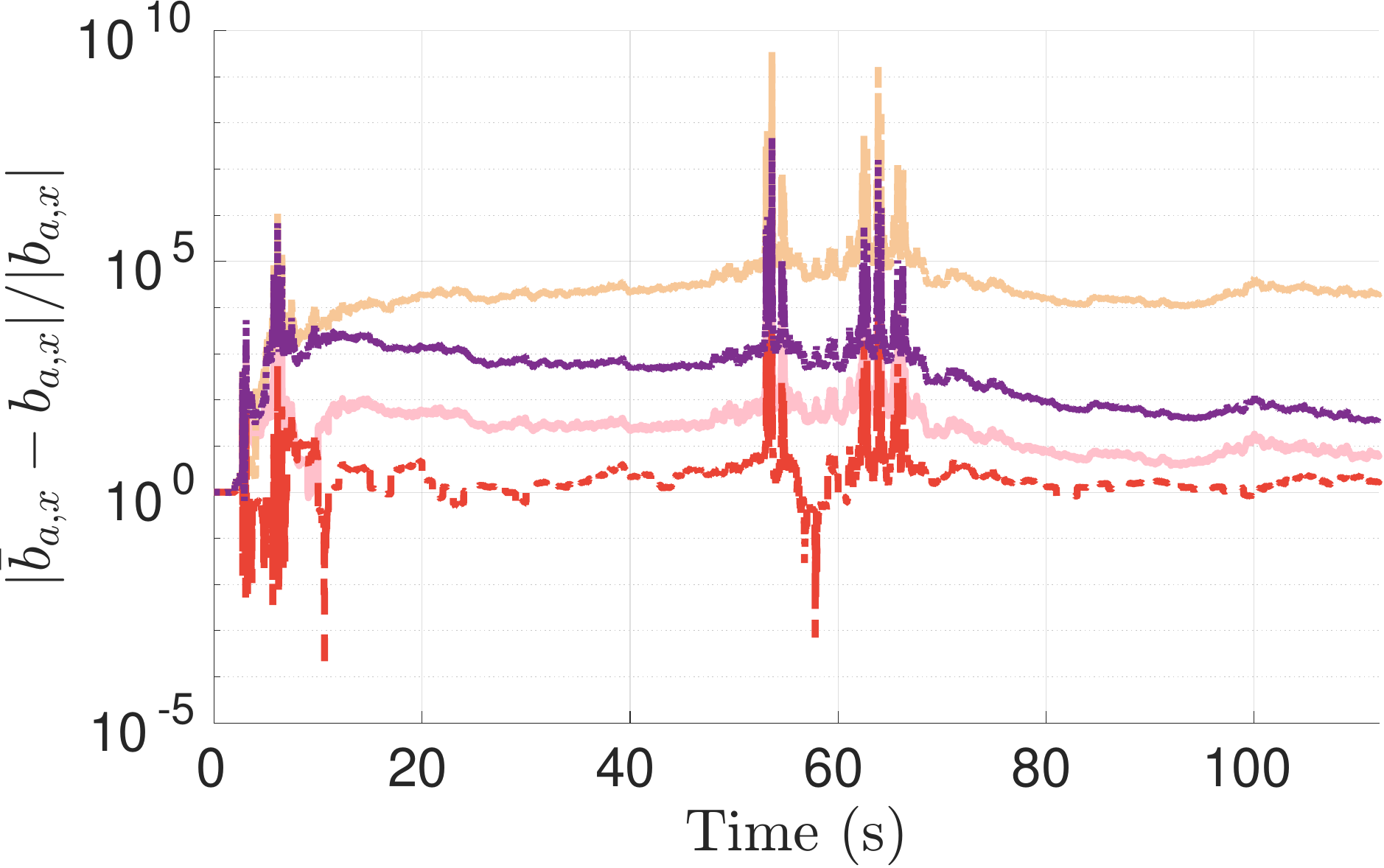}
    }
    \hfill
    \subfloat[\protect\label{fig:accel_bias_y}]{
        \includegraphics[width=0.48\textwidth]{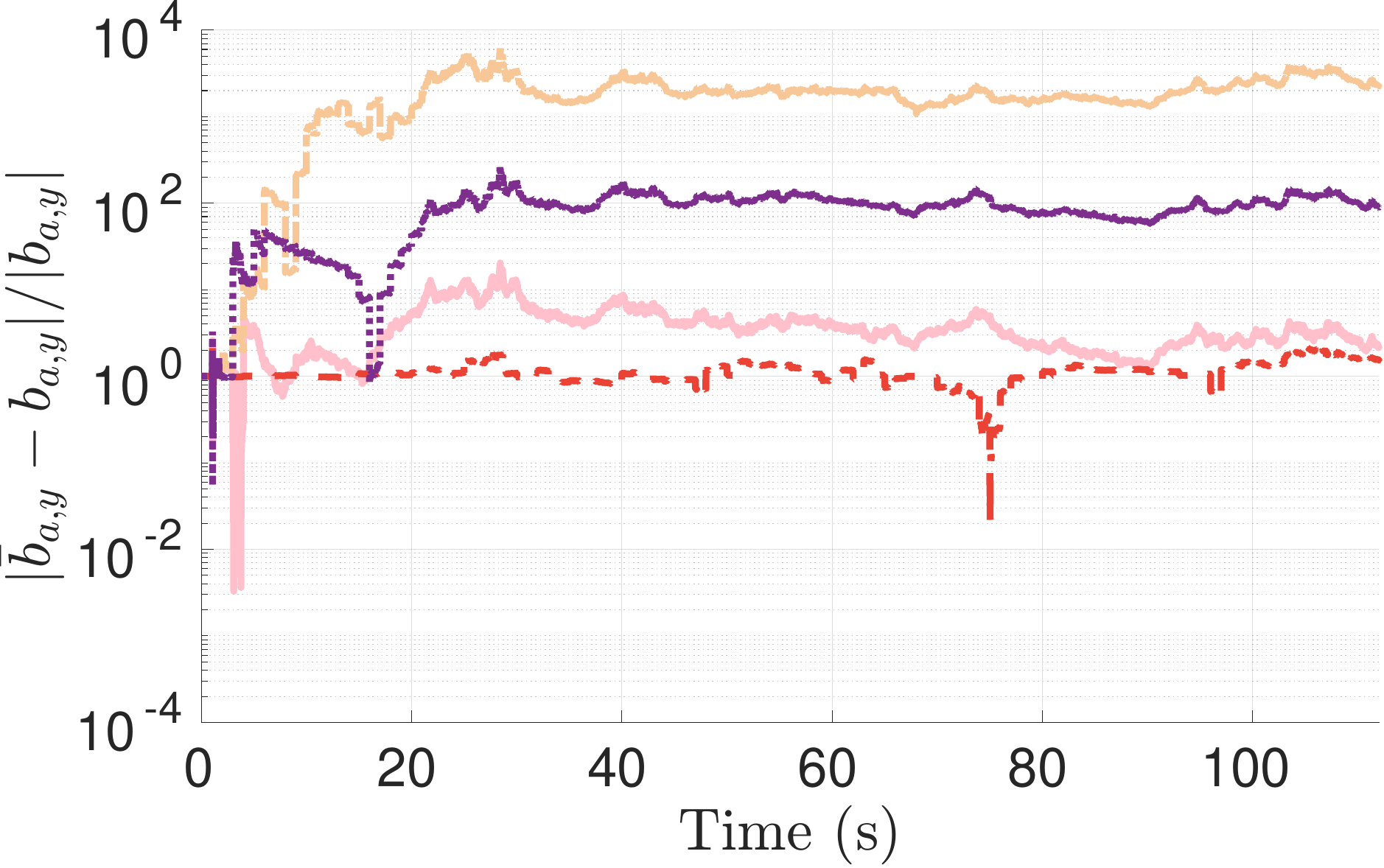}
    }
    \subfloat[\protect\label{fig:accel_bias_z}]{
        \includegraphics[width=0.48\textwidth]{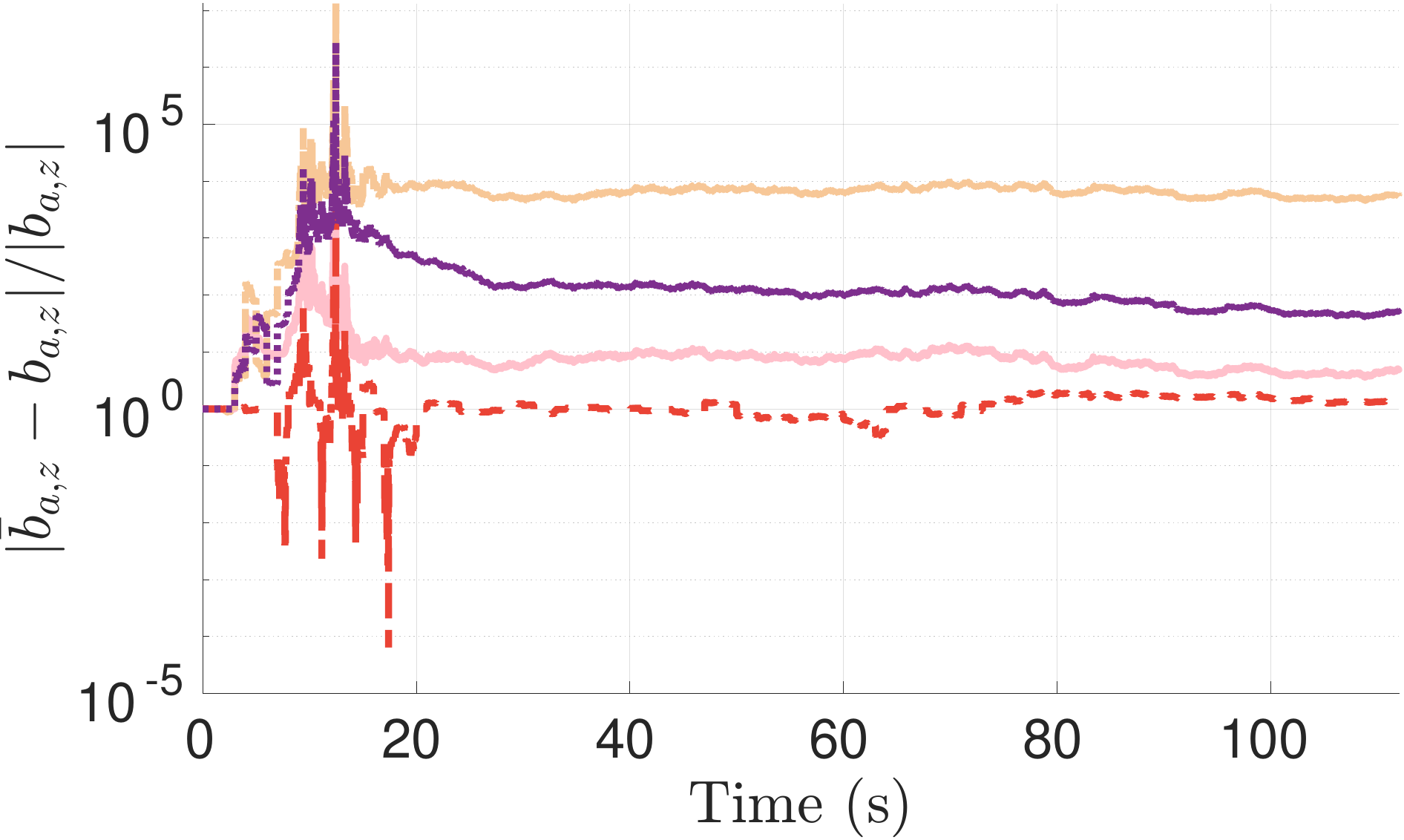}
    }
    \caption{
       Relative error for the estimates of the IMU bias obtained with different filters on one realization of the Monte Carlo simulation. 
       a) Gyroscope bias in the $x$-axis. b) Gyroscope bias in the $y$-axis. c) Gyroscope bias in the $z$-axis. d) Accelerometer bias in the $x$-axis. e) Accelerometer bias in the $y$-axis. f) Accelerometer bias in the $z$-axis.
    }
    \label{fig:bias_estimate}
\end{figure}

The relative estimation errors of the IMU bias for a representative single realization of the Monte Carlo simulation are illustrated in Fig.~\ref{fig:bias_estimate}. Overall, a similar trend to the previous state trajectory evaluations is observed, with the proposed $\text{IterIEKF}$ demonstrating superior convergence and lower residual errors compared to all baseline filters. Moreover, it can be observed in Figs.~\ref{fig:gyro_bias_y} and~\ref{fig:gyro_bias_z} that the invariant filters presented low steady-state error for the gyroscope biases.   
However, every evaluated filter, including the $\text{IterIEKF}$, experiences noticeable difficulty in accurately identifying the accelerometer biases, with the relative error persistently remaining above $1.0$. 
This limitation underscores a core structural constraint of the current formulation: appending the sensor biases directly to the state space as a standard vector addition ($\mathbb{R}^6$) breaks the strict group-affine property of the system dynamics during the propagation step. Consequently, the propagation mechanics are not fully invariant with respect to the bias states. This observability and consistency mismatch could potentially be resolved by adopting the recently introduced Two-Frame Group (TFG) framework~\cite{barrau2022}, a novel Lie group structure designed to embed accelerometer biases in a structurally invariant manner.
\section{Conclusion}
\label{sec:conclusion}

In this work, we have presented the first formulation and systematic evaluation of the Iterated Invariant Extended Kalman Filter tailored for 3D landmark-aided inertial navigation in MAVs. By parameterizing the system state error via right-invariant group actions on $\mathrm{SE}_2(3)$, our framework structurally eliminates the false observability problem that plagues classical non-invariant architectures like the $\mathrm{SO}(3)$-EKF and Iter$\mathrm{SO}(3)$-EKF. Furthermore, by incorporating localized Gauss-Newton optimization loops directly into the invariant measurement update, the IterIEKF overcomes the linearization limitations of the standard single-step IEKF. Extensive Monte Carlo simulations based on a trajectory profile from the EuRoC MAV dataset demonstrate that the IterIEKF yields a profound performance leap, securing upwards of an $80\%$ reduction in both position and linear velocity errors compared to the standard IEKF, while establishing superior NEES statistical consistency and highly accurate gravity vector alignment. While the IterIEKF demonstrates exceptional tracking robustness and prevents cross-state error corruption, online accelerometer biases remain poorly estimated.
Future work aims to scale the proposed filter into a full SLAM framework that unifies invariant state estimation with online data association and unknown landmark map generation.

%
%
%
%
%
\bibliographystyle{splncs04}
\bibliography{main.bib}

@article{santana2026,
  title={Iterated Invariant {EKF} for Quadruped Robot Odometry}, 
  author={Santana, Hilton Marques Souza and Soares, Jo{\~a}o Carlos Virgolino and Goffin, Sven and Nistic{\`o}, Ylenia and Bonnabel, Silv{\`e}re and Semini, Claudio and Meggiolaro, Marco Antonio},
  journal={arXiv preprint arXiv:2604.15449},
  year={2026}
}

@article{goffin2026,
  author={Goffin, Sven and Barrau, Axel and Bonnabel, Silv{\`e}re and Br{\"u}ls, Olivier and Sacr{\`e}, Pierre},
  journal={IEEE Trans. Autom. Control}, 
  title={Iterated Invariant Extended {Kalman} Filter ({IterIEKF})}, 
  year={2026},
  volume={71},
  number={5},
  pages={3380--3387}
}

@INPROCEEDINGS{santana2024,
  author={Santana, Hilton Marques Souza and Soares, João Carlos Virgolino and Nisticò, Ylenia and Meggiolaro, Marco Antonio and Semini, Claudio},
  booktitle={2024 IEEE-RAS 23rd International Conference on Humanoid Robots (Humanoids)}, 
  title={Proprioceptive State Estimation for Quadruped Robots using Invariant Kalman Filtering and Scale-Variant Robust Cost Functions}, 
  year={2024},
  volume={},
  number={},
  pages={213-220}
  }

@article{edward1971,
  author={Edwards Jr., Andrew},
  title={The State of Strapdown Inertial Guidance and Navigation},
  journal={Navigation},
  volume={18},
  number={4},
  pages={386--401},
  year={1971}
}

@article{elhousni2022,
  title={A Survey on Visual Map Localization Using {LiDARs} and Cameras},
  author={Elhousni, Mahdi and Huang, Xinming},
  journal={arXiv preprint arXiv:2208.03376},
  year={2022}
}

@article{debeunne2020,
  title={A review of visual-{LiDAR} fusion based simultaneous localization and mapping},
  author={Debeunne, C{\'e}sar and Vivet, Damien},
  journal={Sensors},
  volume={20},
  number={7},
  pages={2068},
  year={2020}
}

@article{dissanayake2001,
  title={A solution to the simultaneous localization and map building ({SLAM}) problem},
  author={Dissanayake, M. W. M. G. and Newman, Paul and Clark, Steve and Durrant-Whyte, Hugh F. and Csorba, Michael},
  journal={IEEE Trans. Robot. Autom.},
  volume={17},
  number={3},
  pages={229--241},
  year={2001}
}

@book{barfoot2024,
  title={State Estimation for Robotics},
  author={Barfoot, Timothy D.},
  year={2024},
  publisher={Cambridge Univ. Press}
}

@article{leonard1991,
  title={Mobile robot localization by tracking geometric beacons},
  author={Leonard, John J. and Durrant-Whyte, Hugh F. and others},
  journal={IEEE Trans. Robot. Autom.},
  volume={7},
  number={3},
  pages={376--382},
  year={1991}
}

@article{burri2016,
  title={The {EuRoC} micro aerial vehicle datasets},
  author={Burri, Michael and Nikolic, Janosch and Gohl, Pascal and Schneider, Thomas and Rehder, Joern and Omari, Sammy and Achtelik, Markus W. and Siegwart, Roland},
  journal={Int. J. Robot. Res.},
  volume={35},
  number={10},
  pages={1157--1163},
  year={2016}
}

@article{trawny2007,
  title={Vision-aided inertial navigation for pin-point landing using observations of mapped landmarks},
  author={Trawny, Nikolas and Mourikis, Anastasios I. and Roumeliotis, Stergios I. and Johnson, Andrew E. and Montgomery, James F.},
  journal={J. Field Robot.},
  volume={24},
  number={5},
  pages={357--378},
  year={2007}
}

@article{li2013,
  title={High-precision, consistent {EKF}-based visual-inertial odometry},
  author={Li, Mingyang and Mourikis, Anastasios I.},
  journal={Int. J. Robot. Res.},
  volume={32},
  number={6},
  pages={690--711},
  year={2013}
}

@article{barrau2016,
  title={The invariant extended {Kalman} filter as a stable observer},
  author={Barrau, Axel and Bonnabel, Silv{\`e}re},
  journal={IEEE Trans. Autom. Control},
  volume={62},
  number={4},
  pages={1797--1812},
  year={2016}
}

@inproceedings{brossard2018,
  title={Invariant {Kalman} filtering for visual inertial {SLAM}},
  author={Brossard, Martin and Bonnabel, Silv{\`e}re and Barrau, Axel},
  booktitle={Proc. Int. Conf. Inf. Fusion (FUSION)},
  pages={2021--2028},
  year={2018}
}

@article{bloesch2017,
  title={Iterated extended {Kalman} filter based visual-inertial odometry using direct photometric feedback},
   author={Bloesch, Michael and Burri, Michael and Omari, Sammy and Hutter, Marco and Siegwart, Roland},
  journal={Int. J. Robot. Res.},
  volume={36},
  number={10},
  pages={1053--1072},
  year={2017}
}

@article{he2021,
  title={Kalman filters on differentiable manifolds},
  author={He, D. and Xu, W. and Zhang, F.},
  journal={arXiv preprint arXiv:2102.03804},
  year={2021}
}

@article{markley2003,
  title={Multiplicative vs. Additive Filtering for Spacecraft Attitude Determination Quaternion estimation},
  author={Markley, F. Landis},
  journal={J. Guid. Control Dyn.},
  volume={26},
  number={2},
  pages={311--317},
  year={2003}
}

@article{barrau2022,
  title={The geometry of navigation problems},
  author={Barrau, Axel and Bonnabel, Silv{\`e}re},
  journal={IEEE Trans. Autom. Control},
  volume={68},
  number={2},
  pages={689--704},
  year={2022}
}

@article{sola2018,
  title={A micro {Lie} theory for state estimation in robotics},
  author={Sol{\`a}, Joan and Deray, Jeremie and Atchuthan, Dinesh},
  journal={arXiv preprint arXiv:1812.01537},
  year={2018}
}

@article{sola2017,
  title={Quaternion kinematics for the error-state {Kalman} filter},
  author={Sol{\`a}, Joan},
  journal={arXiv preprint arXiv:1711.02508},
  year={2017}
}

@article{brossard2022,
  title={Associating uncertainty to extended poses for on {Lie} group {IMU} preintegration with rotating earth},
  author={Brossard, Martin and Barrau, Axel and Chauchat, Paul and Bonnabel, Silv{\`e}re},
  journal={IEEE Trans. Robot.},
  volume={38},
  number={2},
  pages={998--1015},
  year={2021}
}

@article{herve1999,
title = {The Lie group of rigid body displacements, a fundamental tool for mechanism design},
journal = {Mechanism and Machine Theory},
volume = {34},
number = {5},
pages = {719-730},
year = {1999},
issn = {0094-114X},
author = {JM Hervé}
}
\end{document}